\documentclass{article}
\usepackage{booktabs,graphicx,xcolor,colortbl}
\usepackage{iclr2027_conference,times}
\usepackage{booktabs,graphicx,xcolor,colortbl}
\usepackage{tabularx}
\definecolor{winredfirst}{RGB}{232,143,143}
\definecolor{winredsecond}{RGB}{242,182,182}
\definecolor{winredthird}{RGB}{249,222,222}
\usepackage{amssymb}
\definecolor{winredfirst}{RGB}{232,143,143}
\definecolor{winredsecond}{RGB}{242,182,182}
\definecolor{winredthird}{RGB}{249,222,222}
\usepackage{dblfloatfix}
\providecommand{\winfirst}[1]{#1}
\providecommand{\winsecond}[1]{#1}
\providecommand{\winthird}[1]{#1}
\usepackage{placeins}

\usepackage{amsmath,amsfonts,bm}

\def\eqref#1{equation~\ref{#1}}

\def\1{\bm{1}}

\DeclareMathAlphabet{\mathsfit}{\encodingdefault}{\sfdefault}{m}{sl}
\SetMathAlphabet{\mathsfit}{bold}{\encodingdefault}{\sfdefault}{bx}{n}

\usepackage{multirow}
\usepackage{hyperref}
\usepackage{url}
\usepackage{xcolor}
\usepackage{booktabs}
\usepackage{pifont}
\usepackage{xcolor}
\usepackage{graphicx}
\usepackage{array}
\usepackage{xcolor}
\usepackage{amsmath}
\usepackage{xcolor}
\usepackage{tcolorbox}
\tcbuselibrary{skins,breakable}

\newtcolorbox{divepromptbox}[2][]{%
  enhanced,
  breakable,
  colback=black!2,
  colframe=black,
  colbacktitle=black,
  coltitle=white,
  fonttitle=\bfseries,
  fontupper=\small\rmfamily,
  title={#2},
  boxrule=0.7pt,
  arc=2mm,
  outer arc=2mm,
  left=3mm,
  right=3mm,
  top=2.5mm,
  bottom=2.5mm,
  toptitle=1.5mm,
  bottomtitle=1.5mm,
  before skip=10pt,
  after skip=10pt,
  #1
}

\definecolor{interventionred}{RGB}{196,70,70}
\definecolor{visual}{RGB}{177,156,171}
\definecolor{in}{RGB}{154,158,177}
\newcolumntype{C}[1]{>{\centering\arraybackslash}m{#1}}
\newcommand{\cmark}{%
  \textcolor{green!60!black}{\bfseries\ding{51}}%
}
\newcommand{\xmark}{%
  \textcolor{red!80!black}{\bfseries\ding{55}}%
}

\title{DIVE: Benchmarking Full-Duplex Interaction with Visual Engagement in Continuous Streams}

\iclrfinalcopy
\author{
Yuhang Chen$^{1}$\thanks{Co-first authors.} \quad
Guangyue Liu$^{1}$\footnotemark[1] \quad
Shilong Liu$^{2}$ \\
$^{1}$Huazhong University of Science and Technology \quad
$^{2}$Princeton University \\
\texttt{thisiscyh@gmail.com}
}

\begin{document}

\maketitle

\begin{abstract}
Recent advances in multimodal large language models (MLLMs) have substantially improved spoken interaction and video understanding.  Existing online and streaming benchmarks primarily assess causal multimodal perception through timestamped question answering, whereas real-world interaction requires models to continuously identify task-relevant multimodal cues and decide whether, when, and how to respond. This leaves open how well current models can sustain real-time interaction throughout an evolving multimodal stream. To address this gap, we introduce \textbf{DIVE}, a benchmark for evaluating full-duplex interaction with visual engagement under continuous and realistic multimodal streams. Our benchmark organizes six tasks into three complementary interaction scenarios. \textbf{(1) Proactive Intervention} evaluates whether models can identify emerging visual or semantic cues and intervene at the appropriate moment. \textbf{(2) Continuous Engagement} evaluates sustained participation through continuous visual tracking and visually grounded query interaction. \textbf{(3) Interaction Coordination} evaluates how models coordinate listening and speaking under overlapping and multi-party interactions. \textbf{DIVE} contains 525 audio/video samples with fine-grained human annotations and precise temporal metadata, covering six task-oriented interaction settings grounded in realistic scenarios. Its evaluation protocol separates response quality,
participation decisions, and response timing. We evaluate representative real-time models using response quality,
window response rates, and reaction latency. Experiments show that leading real-time models achieve sub-second
average response latency, yet their average window response rates remain at or below 53.77\%. Response quality varies substantially across models and tasks, with adapting ongoing responses to newly arriving information
remaining particularly challenging. We hope \textbf{DIVE} facilitates progress toward more natural and capable real-time multimodal interactive systems.

\end{abstract}

\section{Introduction}
Multimodal large language models (MLLMs) have made substantial progress in video understanding and multimodal reasoning. Representative systems such as GPT-4o~\cite{hurst2024gpt}, Gemini~\cite{team2023gemini}, and a growing number of open-source MLLMs have demonstrated strong capabilities in understanding multimodal inputs. Correspondingly, widely used benchmarks such as Video-MME~\cite{fu2025video} and LVBench~\cite{wang2025lvbench} evaluate increasingly diverse and long-context video understanding capabilities, yet they largely follow an offline evaluation paradigm. This setting differs fundamentally from real-world interaction. This limitation has motivated growing interest in streaming multimodal understanding. Recent models are designed to process continuously arriving inputs and determine when a response should be produced. Meanwhile, benchmarks such as StreamingBench~\cite{lin2026streamingbench} and OVO-Bench~\cite{niu2025ovo} assess models in real-time settings. In parallel, another line of research has focused on full-duplex interaction, where perception and generation can overlap in time. Early full-duplex systems were predominantly centered on speech interaction. Moshi~\cite{defossez2024moshi} models dialogue as parallel speech-to-speech streams. Correspondingly, the Full-Duplex-Bench~\cite{lin2025full,lin2026full} series evaluates a range of interactive speech behaviors. More recently, full-duplex interaction has been extended to joint visual and audio perception together with native multimodal output, with TML Interaction Models~\cite{thinkingmachines2026interactionmodels} and MiniCPM-o 4.5~\cite{cui2026minicpm} serving as representative examples. Recent benchmarks such as Omni-DuplexEval~\cite{he2026omni} have also begun to evaluate real-time duplex omni-modal interaction through continuous and proactive response scenarios. These benchmarks increasingly cover proactive response, continuous monitoring, and interruption handling, but often assess these capabilities within isolated response windows rather than over the course of sustained interaction.

\begin{table*}[t]
\centering
\caption{
Comparison of \textbf{DIVE} with representative offline, streaming,
and full-duplex interaction benchmarks.
\textbf{V} and \textbf{A} denote video and audio, respectively.
\textbf{Duplex} indicates concurrent perception and generation, while
\textbf{Video Cue} indicates that task-relevant visual cues explicitly determine
when or how the model should respond during an ongoing interaction.
The six task dimensions correspond to our interaction taxonomy:
\textbf{VTI} denotes Video Timely Intervention,
\textbf{SPI} Speech Proactive Intervention,
\textbf{CQI} Continuous Query Interaction,
\textbf{ORA} Overlapping Response Adaptation,
\textbf{AVAT} Audio-Visual Addressee Tracking,
and \textbf{CAT} Continuous Action Tracking.
}
\label{tab:benchmark_comparison}
\renewcommand{\arraystretch}{1.15}
\setlength{\tabcolsep}{4pt}
\resizebox{\textwidth}{!}{%
\begin{tabular}{l*{9}{c}}
\toprule
\multirow{2}{*}{\textbf{Benchmark}}
& \multirow{2}{*}{\textbf{Modality}}
& \multirow{2}{*}{\textbf{Duplex}}
& \multirow{2}{*}{\shortstack{\textbf{Video}\\\textbf{Cue}}}
& \multicolumn{2}{c}{
    \shortstack{\textbf{Proactive}\\\textbf{Intervention}}
  }
& \multicolumn{1}{c}{
    \shortstack{\textbf{Continuous}\\\textbf{Engagement}}
  }
& \multicolumn{2}{c}{
    \shortstack{\textbf{Interaction}\\\textbf{Coordination}}
  }
& \multicolumn{1}{c}{
    \shortstack{\textbf{Continuous}\\\textbf{Engagement}}
  }
\\
\cmidrule(lr){5-6}
\cmidrule(lr){7-7}
\cmidrule(lr){8-9}
\cmidrule(lr){10-10}
& & & &
\textbf{VTI}
& \textbf{SPI}
& \textbf{CQI}
& \textbf{ORA}
& \textbf{AVAT}
& \textbf{CAT}
\\
\midrule

\multicolumn{10}{l}{\textit{Offline Benchmarks}} \\

MVBench~\cite{li2024mvbench}
& V & \xmark & \xmark
& \xmark & \xmark & \xmark & \xmark & \xmark & \xmark \\

Video-MME~\cite{fu2025video}
& V & \xmark & \xmark
& \xmark & \xmark & \xmark & \xmark & \xmark & \xmark \\

MLVU~\cite{zhou2025mlvu}
& V & \xmark & \xmark
& \xmark & \xmark & \xmark & \xmark & \xmark & \xmark \\

LongVideoBench~\cite{wu2024longvideobench}
& V & \xmark & \xmark
& \xmark & \xmark & \xmark & \xmark & \xmark & \xmark \\

WorldSense~\cite{hong2026worldsense}
& V, A & \xmark & \xmark
& \xmark & \xmark & \xmark & \xmark & \xmark & \xmark \\

\midrule
\multicolumn{10}{l}{\textit{Online / Streaming Benchmarks}} \\

StreamingBench~\cite{lin2026streamingbench}
& V, A & \xmark & \cmark
& \cmark & \xmark & \cmark & \xmark & \xmark & \xmark \\

OVO-Bench~\cite{niu2025ovo}
& V & \xmark & \cmark
& \xmark & \xmark & \xmark & \xmark & \xmark & \cmark \\

OmniMMI~\cite{wang2025omnimmi}
& V, A & \xmark & \cmark
& \cmark & \xmark & \cmark & \xmark & \xmark & \xmark \\

ProactiveVideoQA~\cite{wang2025proactivevideoqa}
& V, A & \xmark & \cmark
& \cmark & \xmark & \xmark & \xmark & \xmark & \xmark \\

PhoStream~\cite{lu2026phostream}
& V, A & \xmark & \cmark
& \xmark & \xmark & \xmark & \xmark & \xmark & \xmark \\

RIVER~\cite{shi2026river}
& V & \xmark & \cmark
& \xmark & \xmark & \xmark & \xmark & \xmark & \xmark \\

\midrule
\multicolumn{10}{l}{\textit{Full-Duplex Interaction Benchmarks}} \\

FD-Bench~\cite{lin2025full,lin2026full}
& A & \cmark & \xmark
& \xmark & \xmark & \xmark & \cmark & \xmark & \xmark \\

FLEXI~\cite{ge2025flexi}
& A & \cmark & \xmark
& \xmark & \cmark & \xmark & \cmark & \xmark & \xmark \\

HumDial-FDBench~\cite{wang2026full}
& A & \cmark & \xmark
& \xmark & \xmark & \xmark & \cmark & \xmark & \xmark \\

VideoFDB~\cite{mazumdar2026videofdb}
& V, A & \cmark & \cmark
& \xmark & \xmark & \xmark & \cmark & \xmark & \xmark \\

Omni-DuplexEval~\cite{he2026omni}
& V, A & \cmark & \cmark
& \cmark & \xmark & \xmark & \xmark & \xmark & \xmark \\

OmniInteract~\cite{lu2026omniinteract}
& V, A & \cmark & \cmark
& \cmark & \xmark & \cmark & \cmark & \xmark & \cmark \\

\midrule
\textbf{DIVE}
& \textbf{V, A} & \cmark & \cmark
& \cmark & \cmark & \cmark & \cmark & \cmark & \cmark \\

\bottomrule
\end{tabular}%
}
\par
\setlength{\abovecaptionskip}{4pt}
\setlength{\belowcaptionskip}{4pt}

\caption{
Task-wise distribution of DIVE evaluation instances.
ORA and AVAT each contain 56 base scenarios with two
conditions per scenario, yielding 112 instances per task.
In total, DIVE comprises 525 evaluation instances
constructed from 413 base scenarios.
}
\label{tab:dataset_statistics}

\centering
\small
\renewcommand{\arraystretch}{1.15}
\setlength{\tabcolsep}{8pt}
\begin{tabular}{lrrrrrrr}
\toprule
\textbf{Task}
& \textbf{VTI}
& \textbf{SPI}
& \textbf{CQI}
& \textbf{ORA}
& \textbf{AVAT}
& \textbf{CAT}
& \textbf{Total} \\
\midrule
Instances
& 164 & 50 & 50 & 112 & 112 & 37 & \textbf{525} \\
\bottomrule
\end{tabular}
\end{table*}

To study these interaction requirements, we introduce \textbf{DIVE}, a benchmark for full-duplex audio-visual interaction under continuous and realistic multimodal streams. Our benchmark systematically evaluates three complementary capabilities. \textbf{DIVE} contains 525 audio and video samples with fine-grained annotations and precise temporal information, covering diverse task-oriented scenarios grounded in realistic interactions. Its coverage of both audio and visual settings enables the evaluation of native full-duplex models as well as modality-specific streaming systems. Our experiments reveal that fast responses alone do not ensure reliable full-duplex interaction. Although leading real-time systems achieve sub-second average response latency, their window response rates leave considerable room for improvement, particularly for speech-driven proactive intervention.
Meanwhile, response quality varies substantially across models
and tasks, and adapting ongoing responses to newly arriving
information remains challenging. 
These findings highlight the importance of evaluating response
timing and content quality together under continuous multimodal streams.

\section{Related Works}

\subsection{Duplex Multimodal Models}

Duplex modeling has been explored most extensively in real-time speech interaction. The GPT-Realtime series~\cite{gptrealtime2} and Gemini 3.1 Flash Live~\cite{gemini31live} provide low-latency end-to-end speech interaction. Moshi~\cite{defossez2024moshi} and MoshiRAG~\cite{chien2026moshirag} support simultaneous listening and speaking, and OmniFlatten~\cite{zhang2025omniflatten} enables seamless full-duplex spoken dialogue within a unified autoregressive framework. Related efforts extend real-time speech modeling in complementary directions: MoshiVis~\cite{royer2025visionspeechmodelsteachingspeech} augments duplex speech interaction with visual context, while Voxtral Realtime~\cite{liu2026voxtral} develops stream-synchronous modeling for low-latency speech recognition. More recent systems move toward multimodal duplex interaction. Proprietary systems such as TML Interaction Models~\cite{thinkingmachines2026interactionmodels} and SeedRealtime~\cite{Seeddance2026SeedRealtime} advance real-time interaction to full-duplex multimodal settings, continuously processing audio, video, and text inputs while producing concurrent outputs over ongoing interaction streams. In parallel, openly released models such as DuplexOmni~\cite{huang2026duplexomni} and Aero Realtime~\cite{zhang2026aero} pursue the same paradigm. Together, these systems represent a shift from turn-based multimodal interaction toward continuous full-duplex modeling, where incoming multimodal observations and model responses are jointly coordinated throughout the interaction.

\subsection{Streaming and Duplex Benchmarks}
Recent benchmarks have increasingly evaluated real-time and full-duplex interaction beyond conventional offline understanding. Full-Duplex-Bench~\cite{lin2025full,lin2026full}, FLEXI~\cite{ge2025flexi}, HumDial-FDBench~\cite{wang2026full}, EchoChain~\cite{modi2026echochain}, DuplexSLA~\cite{zhang2026duplexsla}, and DuplexWorld~\cite{bhosale2026duplexworld} primarily focus on spoken interaction, covering turn-taking, interruption, conversational state tracking, and response timing. More recently, evaluation has expanded toward multimodal settings. VideoFDB~\cite{mazumdar2026videofdb} evaluates audio-visual conversational capabilities, while Omni-DuplexEval~\cite{he2026omni} and OmniInteract~\cite{lu2026omniinteract} further broaden the scope of evaluation to include proactive response, continuous monitoring, and interruption-aware interaction over streaming multimodal inputs. In contrast, DIVE complements these settings with explicit task-level requirements for risk intervention, response revision, repeated action counting, and addressee-dependent participation. 
\section{DIVE}

\begin{figure*}[t]
    \centering
    \includegraphics[width=0.83\textwidth]{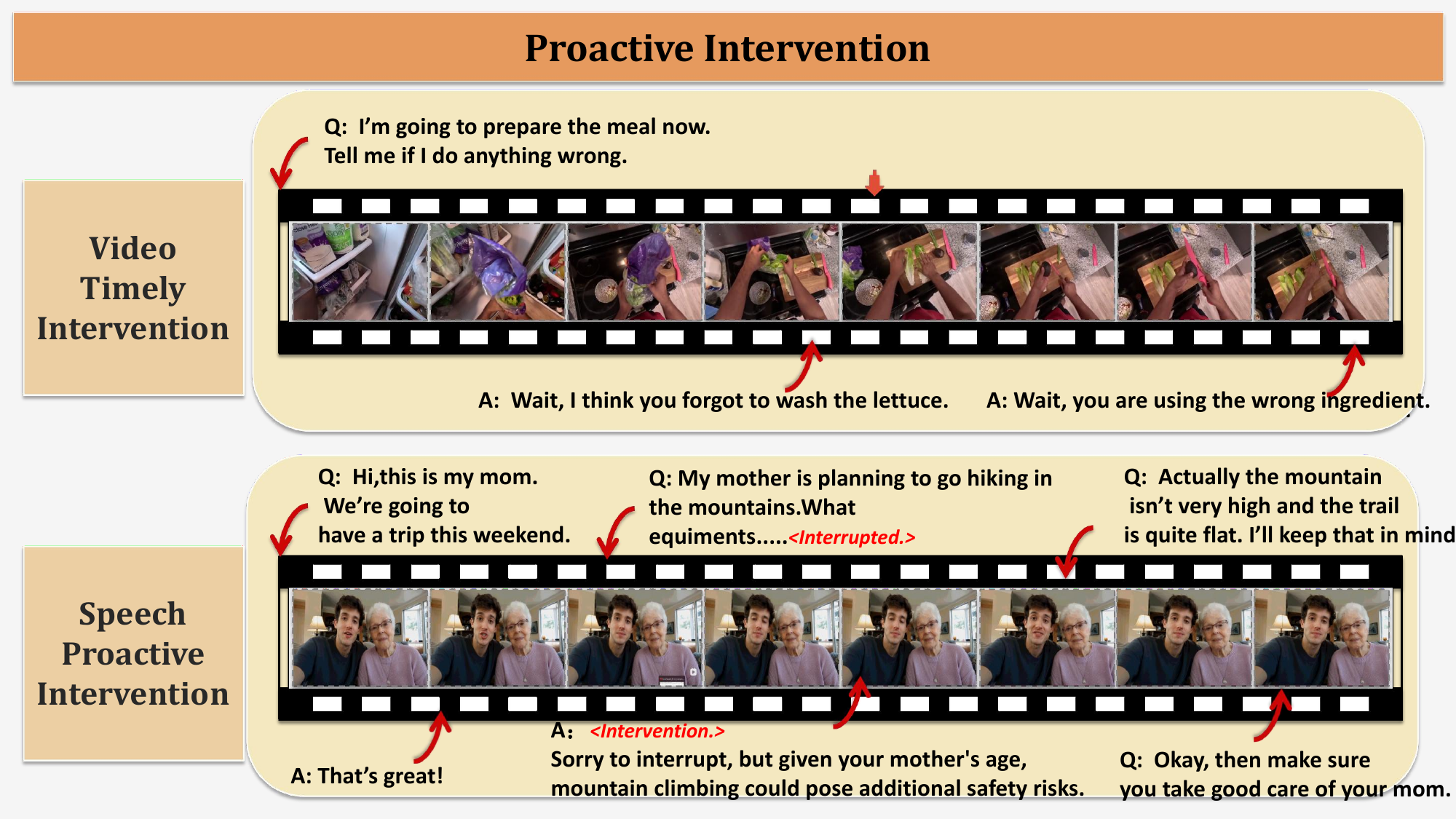}
    \caption{
    Examples of the tasks in Proactive Intervention.
    \textcolor{interventionred}{\textless Interrupted.\textgreater} marks where the ongoing utterance is interrupted,
    whereas \textcolor{interventionred}{\textless Intervention.\textgreater} marks where the model proactively intervenes.
    }
    \label{fig:data1}
    \vspace{-0.8em}
\end{figure*}

\begin{figure*}[t]
    \centering
    \includegraphics[width=0.83\textwidth]{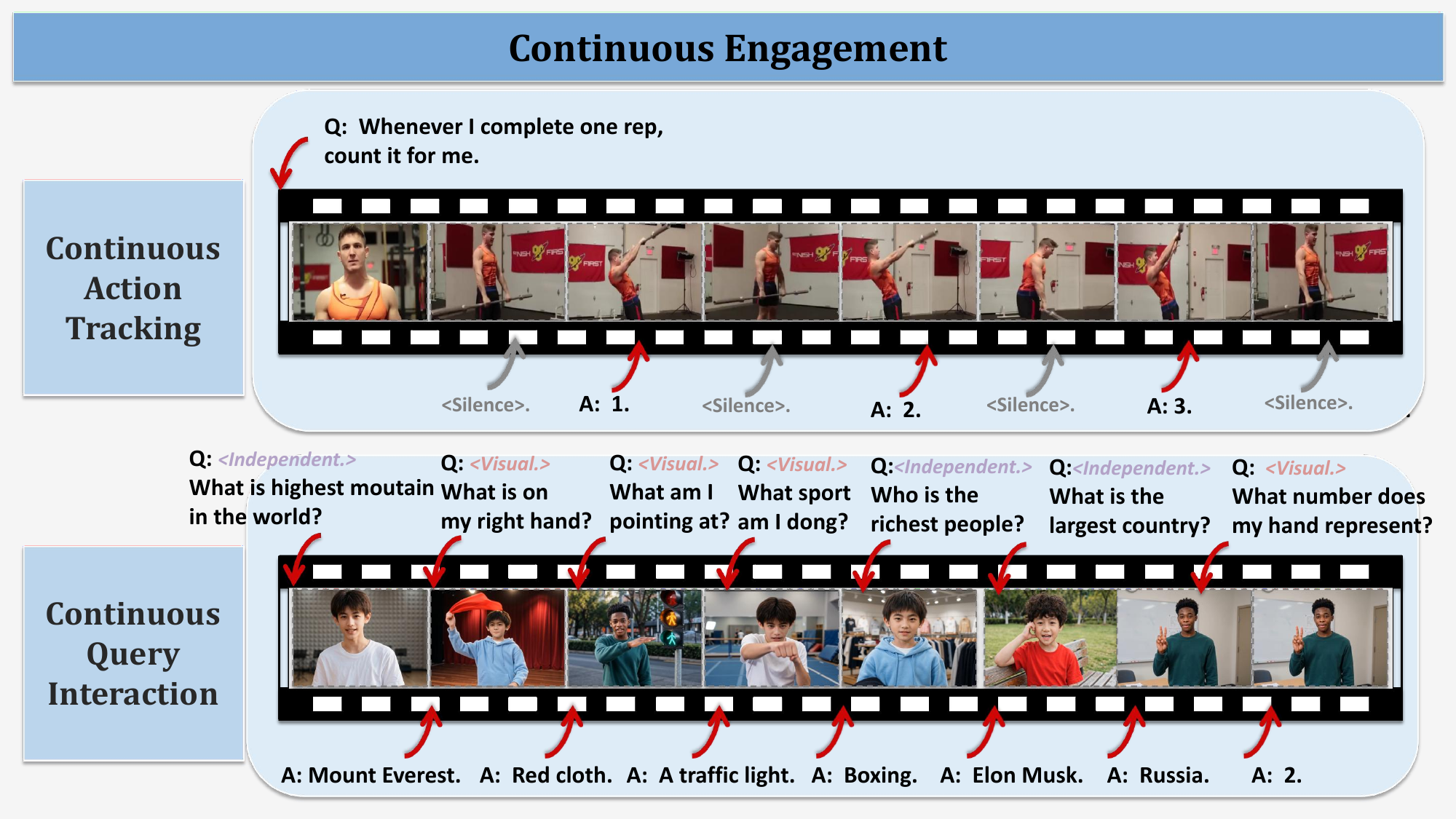}
    \caption{
Examples of the tasks in Continuous Engagement.
\textcolor{visual}{\textless Visual.\textgreater} indicates visually grounded queries,
while \textcolor{in}{\textless Independent.\textgreater} indicates general queries independent of the visual content.
}
    \label{fig:data2}
    \vspace{-0.8em}
\end{figure*}

\subsection{Taxonomy}
\vspace{-0.5em}
Real-time full-duplex interaction requires a model to maintain an evolving understanding of the audio inputs and visual cues, identify task-relevant evidence as it emerges, and integrate it with ongoing
speech to decide how to behave in real time. DIVE therefore organizes interaction
into three complementary scenarios.
\textbf{Proactive Intervention} evaluates whether models can recognize emerging
visual or semantic evidence that calls for timely intervention.
\textbf{Continuous Engagement} evaluates whether models can maintain visual
context over time to track recurring actions and answer successive, visually
grounded queries.
\textbf{Interaction Coordination} evaluates whether models can use ongoing
audio-visual cues to adapt responses under overlap and determine when an
utterance is directed toward the model.
Together, these scenarios cover six tasks that examine how models perceive,
reason over, and act upon continually evolving multimodal context.
DIVE comprises 525 evaluation instances across six tasks,
constructed from 413 base scenarios
(Table~\ref{tab:dataset_statistics}).
ORA and AVAT each contain 56 scenarios with two condition-specific variants, yielding 112 instances per task. The other four tasks contribute 301 instances in total, with one instance per scenario.
\vspace{-0.5em}
\subsection{Proactive Intervention}

Proactive Intervention evaluates whether a model can identify emerging mistakes
or risks from an ongoing interaction and intervene at an appropriate moment.
We consider two complementary settings with visual- and speech-driven triggers.

\noindent\textbf{Video Timely Intervention.} We construct this task from egocentric procedural datasets that provide temporally localized annotations of execution and procedural errors across diverse activities, including PIE-V~\cite{loginova2026correctly}, MATT~\cite{li2026mistake}, Ego-EXTRA~\cite{ragusa2026ego}, and CaptainCook4D~\cite{peddi2024captaincook4d}. Specifically, the model receives a persistent instruction before the activity begins and is expected
to intervene once sufficient visual evidence of the mistake becomes available.
We further annotate the earliest moment at which the mistake becomes visually
identifiable, providing a reference point for evaluating intervention timing.

\noindent\textbf{Speech Proactive Intervention.} We construct speech-driven proactive intervention scenarios from LLaVAShield~\cite{huang2026llavashield} and LlamaPIE~\cite{chen2025llamapie}, supplemented with manually authored cases. We convert unsafe situations into natural spoken narratives and generate corresponding videos with Wan~3.0~\cite{wan2025wan} for scenarios containing visual cues. Risk information is progressively revealed through audio-visual cues, requiring timely intervention once sufficient evidence emerges.

\vspace{-0.8em}
\subsection{Continuous Engagement}
\vspace{-0.8em}
Continuous Engagement evaluates whether a model can maintain active participation in an
evolving interaction and produce repeated or successive responses over time.
We consider two settings involving ongoing action tracking and interleaved
query answering.

\noindent\textbf{Continuous Action Tracking.}
We construct this task from repetitive-action datasets including RepCount~\cite{hu2022transrac},
Multi-RepCount~\cite{zhao2025skim}, and CPR-rac~\cite{liu2026mmaca}.
Their repetition-level annotations are converted into temporally aligned counting
interactions, where the model receives a persistent instruction: it must generate a response upon completion of each target action, and stay silent between repetitions.
This setup evaluates both continuous visual tracking and precise response timing.

\noindent\textbf{Continuous Query Interaction.}
We construct sequential query sequences from Live-Aid~\cite{lei2026live},
supplemented with manually authored scenarios.
Each sequence alternates between two categories of questions: visually grounded questions, whose answers depend on the current video content, and video-independent questions that can be answered without visual evidence.
For manually designed cases, we generate the corresponding videos with
Wan~3.0~\cite{wan2025wan}.
The model must dynamically switch between visual grounding and general
reasoning as successive queries arrive.

\begin{figure*}[t]
    \centering
    \includegraphics[width=0.83\textwidth]{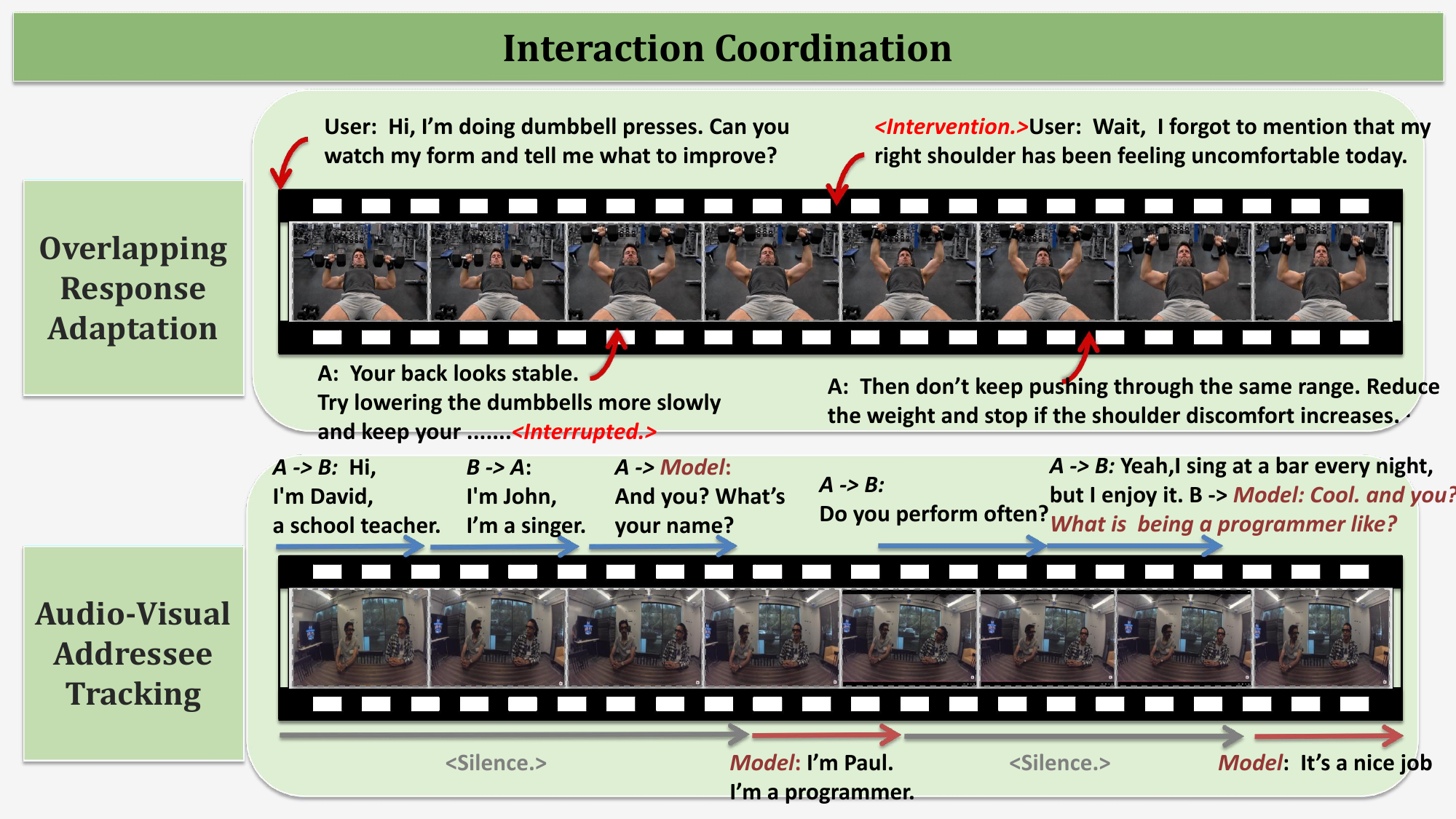}
    \caption{Example of each task in Interaction Coordination. \textcolor{in}{\textless Silence.\textgreater} indicates that the model remains silent.}
    \label{fig:data3}
    \vspace{-0.8em}
\end{figure*}

\subsection{Interaction Coordination}

Interaction Coordination evaluates a model’s ability to produce well‑timed responses that coordinate with incoming user behavior in overlapping and multi‑party interactions. We consider two distinct settings: adapting its responses when interruptions occur, and determining appropriate moments for the model itself to engage in the conversation.

\noindent\textbf{Overlapping Response Adaptation.}
We adapt interruptions and state-update scenarios from EchoChain~\cite{modi2026echochain} and
HumDial-FDBench~\cite{wang2026full}. We rewrite interruptions to introduce
meaningful corrections, new constraints, or contextual updates; we also augment the
original dialogues with visually relevant information. Corresponding videos are
then generated to form audio-visual interactions, requiring the model to continue
perceiving incoming context while speaking and adapt its ongoing response
accordingly.

\begin{figure}[t]
\centering
\includegraphics[width=\linewidth]{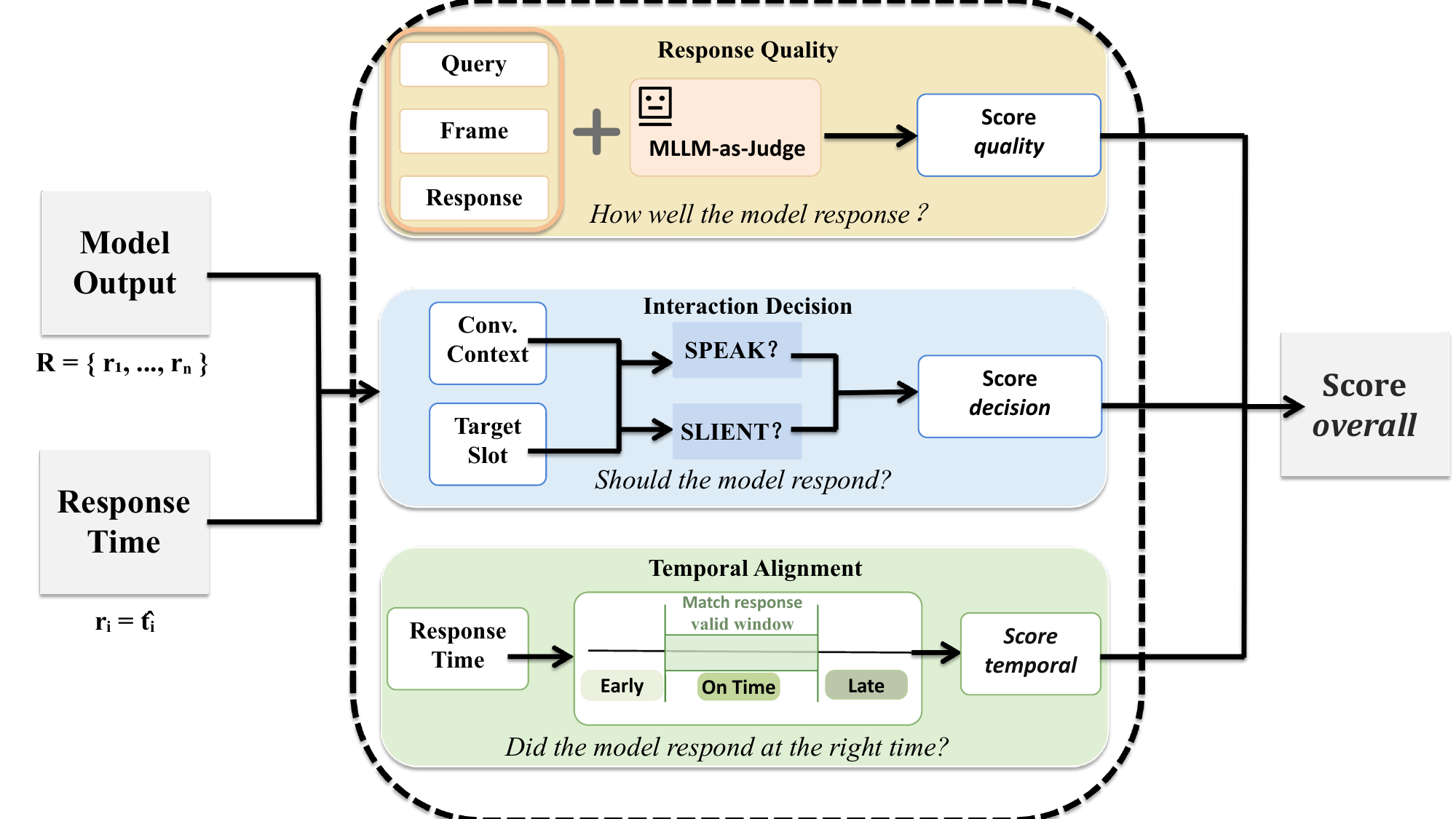}
\caption{
Overview of the DIVE evaluation framework.
\textbf{Interaction Decision}, which measures whether the model
responds when required and remains silent when appropriate;
\textbf{Temporal Alignment}, which categorizes required responses
as early, on-time, late, or missed relative to the annotated
response windows;
and \textbf{Response Quality}, which assesses whether the first
response assigned to each required interaction event satisfies
its task-specific content requirements.
}
\label{fig:evaluation}
\end{figure}
\noindent\textbf{Audio-Visual Addressee Tracking.}
We construct participation patterns from multi-party dialogue resources such as
Speak or Stay Silent~\cite{bhagtani2026speak} and When2Speak~\cite{nama2026when2speak}, where a target
participant must decide whether to speak or remain silent at each interaction
point. We augment these textual conversations with manually crafted visual
cues, and additionally incorporate natural multi-party videos from
Multi-TPC~\cite{lee2026multi} and EasyCom~\cite{donley2021easycom}. The resulting samples require the
model to combine conversational context with cues such as gaze and speaker
orientation to determine whether it is being addressed and should respond.

\subsection{Evaluation}
\label{sec:evaluation}

Conventional multimodal evaluation primarily focuses on whether a model produces a semantically correct response, while continuous full-duplex interaction additionally requires the model to decide whether and when to respond. 
A response may be content-correct yet interactionally inappropriate: the model may speak when it should remain silent, respond prematurely before sufficient evidence is gathered, or reply after the proper interaction window has passed.
We therefore introduce a unified evaluation framework along three complementary dimensions: 
\textbf{Interaction Decision}, \textbf{Temporal Alignment}, and \textbf{Response Quality}. An overview of the evaluation pipeline is shown in Fig.~\ref{fig:evaluation}.
Consider a model output sequence
\begin{equation}
    R = \{r_i=(y_i,\hat{t}_i)\}_{i=1}^{N},
\end{equation}
where $y_i$ denotes the generated response and $\hat{t}_i$ denotes its response timestamp; 
for each annotated interaction slot $j$, DIVE provides the expected interaction decision
$d_j \in \{\textsc{Speak}, \textsc{Silent}\}$ and, when a response is expected, a valid response window
\begin{equation}
W_j=[t_j^{\mathrm{start}},t_j^{\mathrm{end}}).
\end{equation}

\paragraph{Interaction Decision.}
For each task, let $\mathcal{P}$ denote the set of annotated
response-required events. For each $j \in \mathcal{P}$, let
$m_j \in \{0,1\}$ indicate whether the fixed event-matching
procedure assigns a model response to that event, independently
of its content correctness or temporal alignment.
An event without an assigned response has $m_j=0$.
The assigned responses are subsequently evaluated for temporal
alignment and response quality.

\paragraph{Temporal Alignment.}
Temporal Alignment evaluates when the model responds relative
to the availability of sufficient evidence.
For each response-required event $j$, the valid response window
is $W_j=[t_j^{\mathrm{start}},t_j^{\mathrm{end}})$.
Let $\hat{t}_j$ denote the observable onset of its assigned
response on the output channel specified for that model
configuration.
We assign every required event to exactly one temporal category:
\begin{equation}
z_j =
\begin{cases}
\mathrm{Missed},
& m_j=0,\\
\mathrm{Early},
& m_j=1 \;\land\;
  \hat{t}_j<t_j^{\mathrm{start}},\\
\mathrm{On\mbox{-}Time},
& m_j=1 \;\land\;
  t_j^{\mathrm{start}}\leq\hat{t}_j<t_j^{\mathrm{end}},\\
\mathrm{Late},
& m_j=1 \;\land\;
  \hat{t}_j\geq t_j^{\mathrm{end}}.
\end{cases}
\label{eq:temporal_categories}
\end{equation}
All four category rates use $|\mathcal{P}|$ as their denominator
and sum to one.
In particular, the window response rate is
\begin{equation}
\mathrm{WRR}
=
\frac{
\sum_{j \in \mathcal{P}}
[z_j=\mathrm{On\mbox{-}}Time]
}{
|\mathcal{P}|
}.
\label{eq:WRR}
\end{equation}
Content correctness is evaluated separately. We report
$\ell_j=\hat{t}_j-t_j^{\mathrm{start}}$ only for response-required time windows.

\paragraph{Response Quality.}
We use GPT-5.6 Terra to assign a binary content-quality label
$v_{c,k}\in\{0,1\}$ to each model response $k$ assigned to
response-required event $c$ in task $\tau$. Each response is
evaluated against the event-specific instruction, reference requirement, and applicable context. 
Let $\mathcal{C}_{\tau}$ denote the set of annotated response-required events in task $\tau$, 
and let $n_c$ denote the number of responses assigned to event $c$. 
We compute response quality as a response-level
micro-average over all assigned responses:
\begin{equation}
S_{\mathrm{quality}}^{(\tau)}
=
\frac{
\sum_{c\in\mathcal{C}_{\tau}}
\sum_{k=1}^{n_c} v_{c,k}
}{
\sum_{c\in\mathcal{C}_{\tau}} n_c
}.
\label{eq:response_quality}
\end{equation}
Events with no assigned responses do not contribute to this score, while missed responses are evaluated
separately by Interaction Decision and Temporal Alignment.
The same aggregation procedure is applied separately to all six
tasks. 

\section{Experiments}

\begin{table*}[t]
\centering
\caption{
Window response rate (\%) on the six DIVE tasks. Higher is better.
Avg. is the unweighted mean across the six tasks.
}
\label{tab:window_response_rate}
\resizebox{\textwidth}{!}{
\begin{tabular}{lccccccc}
\toprule
\textbf{Model} &
\textbf{VTI $\uparrow$} &
\textbf{SPI $\uparrow$} &
\textbf{CQI $\uparrow$} &
\textbf{ORA $\uparrow$} &
\textbf{AVAT $\uparrow$} &
\textbf{CAT $\uparrow$} &
\textbf{Avg. $\uparrow$} \\
\midrule

\multicolumn{8}{l}{\textit{Native Audio-Visual Models}} \\

Aero-Realtime-4B
& \winthird{53.7}
& 12.3 & 8.7 & 8.2 & 19.9 & 20.3
& 20.52 \\

MiniCPM-o 4.5
& 14.6
& \winfirst{\textbf{24.8}}
& 13.1 & 7.6 & 9.9 & 19.7
& 14.95 \\

DuplexOmni
& 21.1 & 10.6 & 28.6 & 9.1 & 11.8 & 23.4
& 17.43 \\

Qwen3-Omni-30B-A3B-Instruct
& 19.8
& \winthird{20.3}
& 21.2 & 10.1 & 18.5 & 10.0
& 16.65 \\

GPT-Realtime-2.1
& \winfirst{\textbf{100.0}}
& 20.0
& \winfirst{\textbf{44.3}}
& \winfirst{\textbf{56.5}}
& \winfirst{\textbf{41.9}}
& \winfirst{\textbf{59.9}}
& \winfirst{\textbf{53.77}} \\

Gemini-3.1-Flash-Live
& \winsecond{89.6}
& \winsecond{20.9}
& \winsecond{34.2}
& \winsecond{39.7}
& \winthird{25.7}
& \winsecond{33.4}
& \winthird{40.58} \\

Qwen3.5-Omni-Plus-Realtime
& \winfirst{\textbf{100.0}}
& 19.8
& \winthird{31.6}
& \winthird{37.6}
& \winsecond{31.1}
& \winthird{28.5}
& \winsecond{41.43} \\

\midrule

\multicolumn{8}{l}{
\textit{Streaming Visual Models with Timestamp-Aligned Text}
} \\

StreamingVLM
& \winsecond{19.4}
& 11.1
& 12.5
& \winthird{16.8}
& \winthird{17.4}
& 21.5
& \winthird{16.45} \\

Mage-VL
& \winfirst{\textbf{46.5}}
& \winthird{14.6}
& \winfirst{\textbf{22.7}}
& \winsecond{18.8}
& \winfirst{\textbf{24.7}}
& \winthird{23.9}
& \winfirst{\textbf{25.20}} \\

JoyAI-VL-Interaction
& \winthird{17.5}
& \winfirst{\textbf{21.9}}
& \winthird{17.8}
& \winfirst{\textbf{34.7}}
& \winsecond{23.3}
& \winsecond{26.2}
& \winsecond{23.57} \\

MOSS-VL-Realtime
& 10.6
& \winsecond{15.4}
& \winsecond{18.2}
& 14.5
& 8.5
& \winfirst{\textbf{26.3}}
& 15.58 \\

\bottomrule
\end{tabular}
}
\vspace{1mm}
\end{table*}

We evaluate a set of recent real-time multimodal models with native support for both visual and audio inputs, including DuplexOmni~\cite{huang2026duplexomni}, Aero Realtime~\cite{zhang2026aero}, MiniCPM-o 4.5~\cite{cui2026minicpm}, Qwen3-Omni~\cite{xu2025qwen3}, Qwen3.5-Omni-Plus-Realtime~\cite{qwen35omniblog}, Gemini 3.1 Flash Live~\cite{gemini31flashlive} and GPT‑Realtime‑2.1~\cite{gptrealtime21}. For these models, we preserve the original audio and video streams and follow their native streaming or full-duplex inference protocols, allowing them to continuously perceive the evolving audio-visual context and generate responses in real time.

For streaming visual models without native audio input, including
JoyAI-VL-Interaction~\cite{yao2026joyai}, Mage-VL~\cite{yang2026mage}, MOSS-VL-Realtime~\cite{wang2026moss}, and StreamingVLM~\cite{xu2026streamingvlm},
we retain the video and provide timestamp-aligned transcripts in place
of the original audio. Transcript content is revealed incrementally along
the speech timeline. Results for this input
setting are reported separately from those for native audio-visual models.

\begin{table*}[t]
\centering
\caption{
Response quality on the six DIVE tasks. Scores are scaled by
$\times 100$ for readability (higher is better).
Avg. is the unweighted mean across the six tasks.
}
\label{tab:response_quality}

\resizebox{\textwidth}{!}{%
\begin{tabular}{lccccccc}
\toprule
\textbf{Model} &
\textbf{VTI $\uparrow$} &
\textbf{SPI $\uparrow$} &
\textbf{CQI $\uparrow$} &
\textbf{ORA $\uparrow$} &
\textbf{AVAT $\uparrow$} &
\textbf{CAT $\uparrow$} &
\textbf{Avg. $\uparrow$} \\
\midrule

\multicolumn{8}{l}{\textit{Native Audio-Visual Models}} \\

Aero-Realtime-4B
& 2.6 & 13.4 & 9.8 & 2.3 & 0.9 & 4.8 & 5.63 \\

MiniCPM-o 4.5
& 13.2 & 18.7 & 11.6 & 8.8 & 2.3 & 10.3 & 10.82 \\

DuplexOmni
& 11.2 & 16.5 & 9.5 & 7.3 & 14.2 & 8.7 & 11.23 \\

Qwen3-Omni-30B-A3B-Instruct
& 16.6 & 21.5 & 17.2 & 12.3 & 21.2 & 14.1 & 17.15 \\

GPT-Realtime-2.1
& \textbf{65.3}
& 59.7
& \textbf{43.2}
& \textbf{37.8}
& 71.2
& 34.4
& 51.93 \\

Gemini-3.1-Flash-Live
& 59.6
& \textbf{61.3}
& 42.1
& 33.2
& \textbf{88.4}
& \textbf{53.7}
& \textbf{56.38} \\

Qwen3.5-Omni-Plus-Realtime
& 51.2
& 54.2
& 39.5
& 34.2
& 70.4
& 32.9
& 47.07 \\

\midrule

\multicolumn{8}{l}{\textit{Streaming Visual Models with Timestamp-Aligned Text}} \\

StreamingVLM
& 10.7
& 16.4
& \textbf{25.9}
& 11.4
& 25.7
& 23.4
& 18.92 \\

Mage-VL
& 12.8
& 14.9
& 16.8
& 17.3
& \textbf{45.7}
& \textbf{37.9}
& \textbf{24.23} \\

JoyAI-VL-Interaction
& \textbf{27.2}
& \textbf{18.3}
& 17.6
& \textbf{20.1}
& 22.5
& 29.3
& 22.50 \\

MOSS-VL-Realtime
& 23.5
& 10.9
& 11.5
& 14.7
& 18.5
& 19.2
& 16.38 \\

\bottomrule
\end{tabular}%
}

\end{table*}

\subsection{Main Results}
\paragraph{Response Quality Remains Uneven Across Models and Tasks.}
As shown in Table~\ref{tab:response_quality}, Gemini-3.1-Flash-Live
achieves the highest average response quality score of 56.38,
followed by GPT-Realtime-2.1 and Qwen3.5-Omni-Plus-Realtime
with 51.93 and 47.07, respectively.
Among the evaluated open models, Mage-VL obtains the highest
average score of 24.23, indicating a pronounced performance gap between the leading proprietary and open models.
The leading models also exhibit different strengths:
Gemini performs best on SPI, AVAT, and CAT, while GPT-Realtime-2.1 leads on VTI, CQI, and ORA.
Nevertheless, the three highest-scoring models all score lower on ORA than
on VTI and SPI, highlighting the difficulty of adapting response
content when new information arrives during an ongoing response.
These results suggest that current models still struggle to produce consistently high-quality responses across diverse and evolving interaction contexts.

\vspace{-1em}

\paragraph{Low Average Latency Does Not Ensure Consistent Window Compliance. }
Table~\ref{tab:window_response_rate} shows that low average latency
does not necessarily translate into consistently responding within
the valid interaction windows.
GPT-Realtime-2.1 achieves the highest average window response rate
of 53.77\%, followed by Qwen3.5-Omni-Plus-Realtime at 41.43\%
and Gemini-3.1-Flash-Live at 40.58\%.
Despite their sub-second average response latencies,
these models still leave considerable room for improvement
in meeting the annotated response windows.
Performance also varies considerably across tasks.
GPT-Realtime-2.1 and Qwen3.5-Omni-Plus-Realtime both reach
100.0\% on VTI, but achieve only 20.0\% and 19.8\% on SPI,
respectively.
Even the best-performing model on SPI, MiniCPM-o 4.5,
reaches only 24.8\%.
This disparity highlights speech-driven proactive intervention
as a particularly demanding setting, and underscores the need
to evaluate window compliance alongside average response latency.

\vspace{-1em}

\paragraph{Substantial Response Delays Persist in Open Models. }
Proprietary models markedly outperform open models in terms of response latency (Table~\ref{tab:response_latency}). Among all evaluated open models, the best-performing Aero-Realtime-4B still has an average response latency of 2.567 s; several other models take more than 5 s to respond after the earliest appropriate interaction moment. In contrast, recent proprietary real-time interaction systems substantially reduce this delay. Qwen3.5-Omni-Plus-Realtime, GPT-Realtime-2.1, and Gemini 3.1 Flash Live all achieve an average latency below 1 second. This gap suggests that real-time multimodal understanding alone does not guarantee timely interaction, and dedicated real-time interaction systems can significantly improve responsiveness through optimized end-to-end streaming inference.

\vspace{-1em}
\paragraph{Beyond Event Recognition: The Challenge of Timely Proactive Intervention}
DIVE further evaluates whether models can respond promptly once sufficient visual or auditory evidence for intervention becomes available. In the Video Timely Intervention (VTI) task, we measure the temporal offset from the earliest moment at which an error or risk becomes visually identifiable. Open models perform disparately, with response latencies ranging from 1.155 s for Qwen3-Omni to 9.578 s for MOSS-VL-Realtime. These results highlight a challenge beyond conventional streaming understanding: models must not only recognize relevant events, but also determine when the accumulated evidence is sufficient to trigger a proactive response. Timely interaction decisions therefore remain a key bottleneck for current real-time multimodal models.

\section{Limitations}

Although DIVE provides a comprehensive evaluation of continuous audio-visual interaction, several important capabilities fall outside its current scope. First, DIVE does not explicitly evaluate agentic interaction beyond verbal responses. Recent real-time interaction systems have begun to support concurrent tool use and proactive actions during ongoing conversations. In such settings, even without an explicit user request, a model may need to autonomously determine when to retrieve external information, invoke tools, or draw on previously observed context, and then appropriately incorporate the resulting information into the ongoing interaction.

Second, despite DIVE’s emphasis on continuous interaction, the current samples remain limited in duration. Therefore, they do not fully capture challenges that arise over longer time horizons, such as persistent multimodal memory, long-range dependencies, and context accumulation across extended interactions. In addition, controlled within-model comparisons isolating the effects of visual input and concurrent perception are left for future work.

\section{Conclusion}
\begin{table*}[t]
\centering
\caption{
Response latency (seconds) on the six DIVE tasks.
Latency is measured relative to the earliest valid response time,
and lower is better.
}
\label{tab:response_latency}
\resizebox{\textwidth}{!}{
\begin{tabular}{lccccccc}
\toprule
\textbf{Model} &
\textbf{VTI $\downarrow$} &
\textbf{SPI $\downarrow$} &
\textbf{CQI $\downarrow$} &
\textbf{ORA $\downarrow$} &
\textbf{AVAT $\downarrow$} &
\textbf{CAT $\downarrow$} &
\textbf{Avg. $\downarrow$} \\
\midrule

\multicolumn{8}{l}{\textit{Native Audio-Visual Models}} \\

Aero-Realtime-4B
& 2.475 & 2.583 & 1.912 & 3.206 & 3.502 & 1.724 & 2.567 \\

MiniCPM-o 4.5
& 1.811 & \winthird{0.692} & 7.798 & 2.784
& 4.297 & 1.453 & 3.139 \\

DuplexOmni
& 3.108 & 4.775 & 2.110 & 9.918 & 7.380 & 4.772 & 5.344 \\

Qwen3-Omni-30B-A3B-Instruct
& 1.155 & 0.701 & 2.301 & 2.566 & 6.137 & 8.466 & 3.554 \\

GPT-Realtime-2.1
& \winsecond{0.552}
& \winfirst{\textbf{0.413}}
& \winfirst{\textbf{0.395}}
& \winsecond{1.393}
& \winthird{1.671}
& \winfirst{\textbf{0.561}}
& \winsecond{0.831} \\

Gemini-3.1-Flash-Live
& \winthird{0.628}
& 0.799
& \winthird{0.803}
& \winthird{1.858}
& \winfirst{\textbf{0.595}}
& \winsecond{0.591}
& \winthird{0.879} \\

Qwen3.5-Omni-Plus-Realtime
& \winfirst{\textbf{0.487}}
& \winsecond{0.626}
& \winsecond{0.704}
& \winfirst{\textbf{1.113}}
& \winsecond{0.867}
& \winthird{0.613}
& \winfirst{\textbf{0.735}} \\

\midrule

\multicolumn{8}{l}{
\textit{Streaming Visual Models with Timestamp-Aligned Text}
} \\

StreamingVLM
& \winfirst{\textbf{2.042}}
& \winfirst{\textbf{4.632}}
& \winfirst{\textbf{3.119}}
& \winfirst{\textbf{3.138}}
& \winthird{5.864}
& \winsecond{2.983}
& \winfirst{\textbf{3.630}} \\

Mage-VL
& \winthird{8.248}
& 5.689
& \winthird{7.764}
& 8.086
& \winfirst{\textbf{3.519}}
& 9.722
& 7.171 \\

JoyAI-VL-Interaction
& \winsecond{2.949}
& \winthird{5.321}
& \winsecond{3.407}
& \winsecond{3.822}
& 6.692
& \winfirst{\textbf{2.909}}
& \winsecond{4.183} \\

MOSS-VL-Realtime
& 9.578
& \winsecond{4.913}
& 8.247
& \winthird{7.979}
& \winsecond{4.164}
& \winthird{7.405}
& \winthird{7.048} \\

\bottomrule
\end{tabular}
}
\vspace{1mm}
\end{table*}

We introduce DIVE, a benchmark for evaluating full-duplex multimodal interaction with visual engagement under continuous audio-visual streams. With 525 samples covering six tasks across three complementary interaction scenarios, DIVE assesses interaction decisions, temporal alignment, and response quality. Our experiments show that low response latency alone is insufficient for reliable interaction: substantial room for improvement remains in responding within valid windows and consistently producing high-quality responses across tasks. Speech-driven proactive intervention and response adaptation during overlapping interactions continue to pose particular challenges. DIVE provides a testbed for investigating these limitations and advancing multimodal models toward sustained, timely, and contextually appropriate interaction.

\section*{AI Use Statement}

We used large language models to assist with English language
editing, drafting and revising selected manuscript passages,
and literature retrieval and summarization.
AI assistance was also used to prepare LaTeX tables and calculate
summary statistics from experimental results provided by the authors.
As described in the methodology, generative models were used
to construct part of the video data, and an LLM-as-a-Judge
was used to evaluate response quality.
The authors take full responsibility for the accuracy, originality,
and scientific integrity of the final manuscript, including
all AI-assisted text, data, and analyses.
\bibliography{iclr2027_conference}

\begin{thebibliography}{59}
\providecommand{\natexlab}[1]{#1}
\providecommand{\url}[1]{\texttt{#1}}
\expandafter\ifx\csname urlstyle\endcsname\relax
  \providecommand{\doi}[1]{doi: #1}\else
  \providecommand{\doi}{doi: \begingroup \urlstyle{rm}\Url}\fi

\bibitem[Bhagtani et~al.(2026)Bhagtani, Anand, Xu, and Yadav]{bhagtani2026speak}
Kratika Bhagtani, Mrinal Anand, Yu~Chen Xu, and Amit Kumar~Singh Yadav.
\newblock Speak or stay silent: Context-aware turn-taking in multi-party dialogue.
\newblock \emph{arXiv preprint arXiv:2603.11409}, 2026.

\bibitem[Bhosale et~al.(2026)Bhosale, Rajgarhia, Pothanapalli, Shaik, Mukherji, and Manocha]{bhosale2026duplexworld}
Aryan~Vijay Bhosale, Harshit Rajgarhia, Akhil Pothanapalli, Asif Shaik, Abhishek Mukherji, and Dinesh Manocha.
\newblock Duplexworld: Can voice agents help you get through the day?
\newblock \emph{arXiv preprint arXiv:2608.10716}, 2026.

\bibitem[Chen et~al.(2025)Chen, Batchelder, Liu, Smith, and Gollakota]{chen2025llamapie}
Tuochao Chen, Nicholas~Scott Batchelder, Alisa Liu, Noah~A Smith, and Shyamnath Gollakota.
\newblock Llamapie: Proactive in-ear conversation assistants.
\newblock In \emph{Findings of the Association for Computational Linguistics: ACL 2025}, pp.\  13801--13824, 2025.

\bibitem[Chien et~al.(2026)Chien, Orsini, Kharitonov, Zeghidour, Livescu, and D{\'e}fossez]{chien2026moshirag}
Chung-Ming Chien, Manu Orsini, Eugene Kharitonov, Neil Zeghidour, Karen Livescu, and Alexandre D{\'e}fossez.
\newblock Moshirag: Asynchronous knowledge retrieval for full-duplex speech language models.
\newblock \emph{arXiv preprint arXiv:2604.12928}, 2026.

\bibitem[Cui et~al.(2026)Cui, Xu, Wang, Yu, Sun, Xu, Wang, He, Ma, Cai, et~al.]{cui2026minicpm}
Junbo Cui, Bokai Xu, Chongyi Wang, Tianyu Yu, Weiyue Sun, Yingjing Xu, Tianran Wang, Zhihui He, Wenshuo Ma, Tianchi Cai, et~al.
\newblock Minicpm-o 4.5: Towards real-time full-duplex omni-modal interaction.
\newblock \emph{arXiv preprint arXiv:2604.27393}, 2026.

\bibitem[D{\'e}fossez et~al.(2024)D{\'e}fossez, Mazar{\'e}, Orsini, Royer, P{\'e}rez, J{\'e}gou, Grave, and Zeghidour]{defossez2024moshi}
Alexandre D{\'e}fossez, Laurent Mazar{\'e}, Manu Orsini, Am{\'e}lie Royer, Patrick P{\'e}rez, Herv{\'e} J{\'e}gou, Edouard Grave, and Neil Zeghidour.
\newblock Moshi: a speech-text foundation model for real-time dialogue.
\newblock \emph{arXiv preprint arXiv:2410.00037}, 2024.

\bibitem[Donley et~al.(2021)Donley, Tourbabin, Lee, Broyles, Jiang, Shen, Pantic, Ithapu, and Mehra]{donley2021easycom}
Jacob Donley, Vladimir Tourbabin, Jung-Suk Lee, Mark Broyles, Hao Jiang, Jie Shen, Maja Pantic, Vamsi~Krishna Ithapu, and Ravish Mehra.
\newblock Easycom: An augmented reality dataset to support algorithms for easy communication in noisy environments.
\newblock \emph{arXiv preprint arXiv:2107.04174}, 2021.

\bibitem[Fu et~al.(2025)Fu, Dai, Luo, Li, Ren, Zhang, Wang, Zhou, Shen, Zhang, et~al.]{fu2025video}
Chaoyou Fu, Yuhan Dai, Yongdong Luo, Lei Li, Shuhuai Ren, Renrui Zhang, Zihan Wang, Chenyu Zhou, Yunhang Shen, Mengdan Zhang, et~al.
\newblock Video-mme: The first-ever comprehensive evaluation benchmark of multi-modal llms in video analysis.
\newblock In \emph{2025 IEEE/CVF Conference on Computer Vision and Pattern Recognition (CVPR)}, pp.\  24108--24118. IEEE, 2025.

\bibitem[Ge et~al.(2025)Ge, Chen, Xiao, Liu, Xiao, Xiang, Yu, and Zhu]{ge2025flexi}
Yuan Ge, Saihan Chen, Jingqi Xiao, Xiaoqian Liu, Tong Xiao, Yan Xiang, Zhengtao Yu, and Jingbo Zhu.
\newblock Flexi: Benchmarking full-duplex human-llm speech interaction.
\newblock \emph{arXiv preprint arXiv:2509.22243}, 2025.

\bibitem[{Google DeepMind}(2026{\natexlab{a}})]{gemini31flashlive}
{Google DeepMind}.
\newblock Gemini 3.1 flash audio (flash live, tts), March 2026{\natexlab{a}}.
\newblock Model Card.

\bibitem[{Google DeepMind}(2026{\natexlab{b}})]{gemini31live}
{Google DeepMind}.
\newblock Gemini 3.1 flash live: Making audio ai more natural and reliable.
\newblock Google DeepMind Model Card, March 2026{\natexlab{b}}.
\newblock URL \url{https://blog.google/innovation-and-ai/models-and-research/gemini-models/gemini-3-1-flash-live/}.

\bibitem[He et~al.(2026)He, Xiang, Xu, Xu, Cui, Zhou, Yao, and Wen]{he2026omni}
Chaoqun He, Mingyang Xiang, Yingjing Xu, Bokai Xu, Junbo Cui, Jie Zhou, Yuan Yao, and Lijie Wen.
\newblock Omni-duplexeval: Evaluating real-time duplex omni-modal interaction.
\newblock \emph{arXiv preprint arXiv:2605.17360}, 2026.

\bibitem[Hong et~al.(2026)Hong, Yan, Cai, Jiang, Hu, and Xie]{hong2026worldsense}
Jack Hong, Shilin Yan, Jiayin Cai, Xiaolong Jiang, Yao Hu, and Weidi Xie.
\newblock Worldsense: Evaluating real-world omnimodal understanding for multimodal llms.
\newblock In \emph{International Conference on Learning Representations}, volume 2026, pp.\  52423--52443, 2026.

\bibitem[Hu et~al.(2022)Hu, Dong, Zhao, Lian, Li, and Gao]{hu2022transrac}
Huazhang Hu, Sixun Dong, Yiqun Zhao, Dongze Lian, Zhengxin Li, and Shenghua Gao.
\newblock Transrac: Encoding multi-scale temporal correlation with transformers for repetitive action counting.
\newblock In \emph{2022 IEEE/CVF Conference on Computer Vision and Pattern Recognition (CVPR)}, pp.\  18991--19000. IEEE, 2022.

\bibitem[Huang et~al.(2026{\natexlab{a}})Huang, Peng, Xu, Huang, Lu, and Shen]{huang2026llavashield}
Guolei Huang, Qinzhi Peng, Gan Xu, Yao Huang, Yuxuan Lu, and Yongjun Shen.
\newblock Llavashield: Safeguarding multimodal multi-turn dialogues in vision-language models.
\newblock In \emph{Proceedings of the IEEE/CVF Conference on Computer Vision and Pattern Recognition}, pp.\  30130--30140, 2026{\natexlab{a}}.

\bibitem[Huang et~al.(2026{\natexlab{b}})Huang, Zhang, Yu, Shi, Ma, Xu, Gao, Hao, He, and Liu]{huang2026duplexomni}
Muye Huang, Lingling Zhang, Xingyu Yu, Lei Shi, Zhanyu Ma, Jun Xu, Jiuchong Gao, Jinghua Hao, Renqing He, and Jun Liu.
\newblock Duplexomni: Real-time listening, seeing, thinking, and speaking for full-duplex interaction.
\newblock \emph{arXiv preprint arXiv:2606.09186}, 2026{\natexlab{b}}.

\bibitem[Hurst et~al.(2024)Hurst, Lerer, Goucher, Perelman, Ramesh, Clark, Ostrow, Welihinda, Hayes, Radford, et~al.]{hurst2024gpt}
Aaron Hurst, Adam Lerer, Adam~P Goucher, Adam Perelman, Aditya Ramesh, Aidan Clark, AJ~Ostrow, Akila Welihinda, Alan Hayes, Alec Radford, et~al.
\newblock Gpt-4o system card.
\newblock \emph{arXiv preprint arXiv:2410.21276}, 2024.

\bibitem[Lab(2026)]{thinkingmachines2026interactionmodels}
Thinking~Machines Lab.
\newblock Interaction models: A scalable approach to human-ai collaboration.
\newblock \emph{Thinking Machines Lab: Connectionism}, May 2026.
\newblock \doi{10.64434/tml.20260511}.
\newblock https://thinkingmachines.ai/blog/interaction-models/.

\bibitem[Lee \& Deng(2026)Lee and Deng]{lee2026multi}
Meng-Chen Lee and Zhigang Deng.
\newblock Multi-tpc: A multimodal dataset for three-party conversations with speech, motion, and gaze.
\newblock \emph{Scientific Data}, 13\penalty0 (1):\penalty0 429, 2026.

\bibitem[Lei et~al.(2026)Lei, Fan, Liu, Dong, Qiu, Leng, Liu, Chen, Gao, and Wang]{lei2026live}
Yiming Lei, Yize Fan, Zeming Liu, Jiaji Dong, Hui Qiu, Haitao Leng, Qingjie Liu, Kehai Chen, Tingting Gao, and Yunhong Wang.
\newblock Live-aid: A large-scale dialogue dataset and benchmark for interleaved multi-party interactions in live streaming.
\newblock In \emph{Findings of the Association for Computational Linguistics: ACL 2026}, pp.\  23813--23850, 2026.

\bibitem[Li et~al.(2024)Li, Wang, He, Li, Wang, Liu, Wang, Xu, Chen, Luo, et~al.]{li2024mvbench}
Kunchang Li, Yali Wang, Yinan He, Yizhuo Li, Yi~Wang, Yi~Liu, Zun Wang, Jilan Xu, Guo Chen, Ping Luo, et~al.
\newblock Mvbench: A comprehensive multi-modal video understanding benchmark.
\newblock In \emph{Proceedings of the IEEE/CVF conference on computer vision and pattern recognition}, pp.\  22195--22206, 2024.

\bibitem[Li et~al.(2026)Li, Jain, Bellos, and Corso]{li2026mistake}
Yayuan Li, Aadit Jain, Filippos Bellos, and Jason~J Corso.
\newblock Mistake attribution: Fine-grained mistake understanding in egocentric videos.
\newblock In \emph{Proceedings of the IEEE/CVF Conference on Computer Vision and Pattern Recognition}, pp.\  23966--23976, 2026.

\bibitem[Lin et~al.(2025)Lin, Lian, Li, Wang, Anumanchipalli, Liu, and Lee]{lin2025full}
Guan-Ting Lin, Jiachen Lian, Tingle Li, Qirui Wang, Gopala Anumanchipalli, Alexander~H Liu, and Hung-yi Lee.
\newblock Full-duplex-bench: A benchmark to evaluate full-duplex spoken dialogue models on turn-taking capabilities.
\newblock In \emph{2025 IEEE Automatic Speech Recognition and Understanding Workshop (ASRU)}, pp.\  1--8. IEEE, 2025.

\bibitem[Lin et~al.(2026{\natexlab{a}})Lin, Kuan, Wang, Lian, Li, Watanabe, and Lee]{lin2026full}
Guan-Ting Lin, Shih-Yun~Shan Kuan, Qirui Wang, Jiachen Lian, Tingle Li, Shinji Watanabe, and Hung-yi Lee.
\newblock Full-duplex-bench v1. 5: Evaluating overlap handling for full-duplex speech models.
\newblock In \emph{ICASSP 2026-2026 IEEE International Conference on Acoustics, Speech and Signal Processing (ICASSP)}, pp.\  19447--19451. IEEE, 2026{\natexlab{a}}.

\bibitem[Lin et~al.(2026{\natexlab{b}})Lin, Fang, Chen, Cheng, Wan, Luo, Wang, Li, Liu, and Sun]{lin2026streamingbench}
Junming Lin, Zheng Fang, Chi Chen, Haoxuan Cheng, Zihao Wan, Fuwen Luo, Ziyue Wang, Peng Li, Yang Liu, and Maosong Sun.
\newblock Streamingbench: Assessing the gap for mllms to achieve streaming video understanding.
\newblock In \emph{ICASSP 2026-2026 IEEE International Conference on Acoustics, Speech and Signal Processing (ICASSP)}, pp.\  12147--12151. IEEE, 2026{\natexlab{b}}.

\bibitem[Liu et~al.(2026{\natexlab{a}})Liu, Ehrenberg, Lo, Sun, Lample, Delignon, Chandu, von Platen, Muddireddy, Arora, et~al.]{liu2026voxtral}
Alexander~H Liu, Andy Ehrenberg, Andy Lo, Chen-Yo Sun, Guillaume Lample, Jean-Malo Delignon, Khyathi~Raghavi Chandu, Patrick von Platen, Pavankumar~Reddy Muddireddy, Rohin Arora, et~al.
\newblock Voxtral realtime.
\newblock \emph{arXiv preprint arXiv:2602.11298}, 2026{\natexlab{a}}.

\bibitem[Liu et~al.(2026{\natexlab{b}})Liu, Ruan, and Zheng]{liu2026mmaca}
Huaiyang Liu, Jun Ruan, and Haifeng Zheng.
\newblock Mmaca-rac: A multi-stage multimodal fusion framework with auto-correlation attention for repetitive action counting.
\newblock \emph{Computer Vision and Image Understanding}, pp.\  104846, 2026{\natexlab{b}}.

\bibitem[Loginova \& Keller(2026)Loginova and Keller]{loginova2026correctly}
Olga Loginova and Frank Keller.
\newblock How to correctly make mistakes: A framework for constructing and benchmarking mistake aware egocentric procedural videos.
\newblock \emph{arXiv preprint arXiv:2604.15134}, 2026.

\bibitem[Lu et~al.(2026{\natexlab{a}})Lu, Guan, Bo, Chen, Guo, Li, Liu, Sun, Li, Zhang, et~al.]{lu2026phostream}
Xudong Lu, Huankang Guan, Yang Bo, Jinpeng Chen, Xintong Guo, Shuhan Li, Fang Liu, Peiwen Sun, Xueying Li, Wei Zhang, et~al.
\newblock Phostream: Benchmarking real-world streaming for omnimodal assistants in mobile scenarios.
\newblock \emph{arXiv preprint arXiv:2601.22575}, 2026{\natexlab{a}}.

\bibitem[Lu et~al.(2026{\natexlab{b}})Lu, Li, Wang, Bo, Chen, Li, Yang, Liu, Yang, Hou, et~al.]{lu2026omniinteract}
Xudong Lu, Xueying Li, Annan Wang, Yang Bo, Jinpeng Chen, Zengliang Li, Nianzu Yang, Rui Liu, Xue Yang, Jingwen Hou, et~al.
\newblock Omniinteract: Benchmarking real-world streaming interaction for real-time omnimodal assistants.
\newblock \emph{arXiv preprint arXiv:2605.26485}, 2026{\natexlab{b}}.

\bibitem[Mazumdar et~al.(2026)Mazumdar, Park, Roy, Srihari, Wang, Zhou, Wang, Nagano, and De~Mello]{mazumdar2026videofdb}
Amrita Mazumdar, Seonwook Park, Rajarshi Roy, Nikhil Srihari, Shengze Wang, Yuhao Zhou, Julia Wang, Koki Nagano, and Shalini De~Mello.
\newblock Videofdb: Evaluating full-duplex vision-speech capabilities in conversational agents.
\newblock \emph{arXiv preprint arXiv:2605.30256}, 2026.

\bibitem[Modi et~al.(2026)Modi, Mahajan, Wetter, and Welles]{modi2026echochain}
Smit~Nautambhai Modi, Gandharv Mahajan, Marc Wetter, and Randall Welles.
\newblock Echochain: A full-duplex benchmark for state-update reasoning under interruptions.
\newblock \emph{arXiv preprint arXiv:2604.16456}, 2026.

\bibitem[Nama et~al.(2026)Nama, Mendi, Ye, and Bent]{nama2026when2speak}
Vihaan Nama, Shreya Mendi, Zian Ye, and Brinnae Bent.
\newblock When2speak: A dataset for temporal participation and turn-taking in multi-party conversations for large language models.
\newblock \emph{arXiv preprint arXiv:2605.05626}, 2026.

\bibitem[Niu et~al.(2025)Niu, Li, Miao, Ge, Zhou, He, Dong, Duan, Ding, Qian, et~al.]{niu2025ovo}
Junbo Niu, Yifei Li, Ziyang Miao, Chunjiang Ge, Yuanhang Zhou, Qihao He, Xiaoyi Dong, Haodong Duan, Shuangrui Ding, Rui Qian, et~al.
\newblock Ovo-bench: How far is your video-llms from real-world online video understanding?
\newblock In \emph{2025 IEEE/CVF Conference on Computer Vision and Pattern Recognition (CVPR)}, pp.\  18902--18913. IEEE, 2025.

\bibitem[{OpenAI}(2025)]{gptrealtime2}
{OpenAI}.
\newblock Advancing voice intelligence with new models in the api.
\newblock OpenAI, 2025.
\newblock URL \url{https://openai.com/index/introducing-gpt-realtime/}.

\bibitem[{OpenAI}(2026)]{gptrealtime21}
{OpenAI}.
\newblock Gpt-realtime-2.1, 2026.
\newblock URL \url{https://developers.openai.com/api/docs/models/gpt-realtime-2.1}.

\bibitem[Peddi et~al.(2024)Peddi, Arya, Challa, Pallapothula, Vyas, Gouripeddi, Zhang, Wang, Komaragiri, Ragan, et~al.]{peddi2024captaincook4d}
Rohith Peddi, Shivvrat Arya, Bharath Challa, Likhitha Pallapothula, Akshay Vyas, Bhavya Gouripeddi, Qifan Zhang, Jikai Wang, Vasundhara Komaragiri, Eric Ragan, et~al.
\newblock Captaincook4d: A dataset for understanding errors in procedural activities.
\newblock \emph{Advances in Neural Information Processing Systems}, 37:\penalty0 135626--135679, 2024.

\bibitem[Ragusa et~al.(2026)Ragusa, Mazzamuto, Forte, D'Ambra, Fort, Engel, Furnari, and Farinella]{ragusa2026ego}
Francesco Ragusa, Michele Mazzamuto, Rosario Forte, Irene D'Ambra, James Fort, Jakob Engel, Antonino Furnari, and Giovanni~Maria Farinella.
\newblock Ego-extra: video-language egocentric dataset for expert-trainee assistance.
\newblock In \emph{Proceedings of the IEEE/CVF Winter Conference on Applications of Computer Vision}, pp.\  4438--4450, 2026.

\bibitem[Royer et~al.(2025)Royer, Böhle, de~Marmiesse, Mazaré, Zeghidour, Défossez, and Pérez]{royer2025visionspeechmodelsteachingspeech}
Amélie Royer, Moritz Böhle, Gabriel de~Marmiesse, Laurent Mazaré, Neil Zeghidour, Alexandre Défossez, and Patrick Pérez.
\newblock Vision-speech models: Teaching speech models to converse about images, 2025.
\newblock URL \url{https://arxiv.org/abs/2503.15633}.

\bibitem[Seeddance(2026)]{Seeddance2026SeedRealtime}
Seeddance.
\newblock Seedrealtime: An audio-visual full-duplex llm.
\newblock \emph{Seeddance}, August 2026.
\newblock https://seed.bytedance.com/en/SeedRealtime.

\bibitem[Shi et~al.(2026)Shi, Zhao, Jiang, Zeng, Wang, and Wang]{shi2026river}
Yansong Shi, Qingsong Zhao, Tianxiang Jiang, Xiangyu Zeng, Yi~Wang, and Limin Wang.
\newblock River: A real-time interaction benchmark for video llms.
\newblock \emph{arXiv preprint arXiv:2603.03985}, 2026.

\bibitem[Team et~al.(2023)Team, Anil, Borgeaud, Alayrac, Yu, Soricut, Schalkwyk, Dai, Hauth, Millican, et~al.]{team2023gemini}
Gemini Team, Rohan Anil, Sebastian Borgeaud, Jean-Baptiste Alayrac, Jiahui Yu, Radu Soricut, Johan Schalkwyk, Andrew~M Dai, Anja Hauth, Katie Millican, et~al.
\newblock Gemini: a family of highly capable multimodal models.
\newblock \emph{arXiv preprint arXiv:2312.11805}, 2023.

\bibitem[Team(2026)]{qwen35omniblog}
Qwen Team.
\newblock Qwen3.5-omni: Scaling up, toward native omni-modal agi, March 2026.
\newblock URL \url{https://qwen.ai/blog?id=qwen3.5-omni}.

\bibitem[Wan et~al.(2025)Wan, Wang, Ai, Wen, Mao, Xie, Chen, Yu, Zhao, Yang, et~al.]{wan2025wan}
Team Wan, Ang Wang, Baole Ai, Bin Wen, Chaojie Mao, Chen-Wei Xie, Di~Chen, Feiwu Yu, Haiming Zhao, Jianxiao Yang, et~al.
\newblock Wan: Open and advanced large-scale video generative models.
\newblock \emph{arXiv preprint arXiv:2503.20314}, 2025.

\bibitem[Wang et~al.(2026{\natexlab{a}})Wang, Xue, Li, Zhao, Wang, Wang, Xu, Bu, and Xie]{wang2026full}
Chengyou Wang, Hongfei Xue, Guojian Li, Zhixian Zhao, Shuiyuan Wang, Shuai Wang, Xin Xu, Hui Bu, and Lei Xie.
\newblock Full-duplex interaction in spoken dialogue systems: A comprehensive study from the icassp 2026 humdial challenge.
\newblock \emph{arXiv preprint arXiv:2604.21406}, 2026{\natexlab{a}}.

\bibitem[Wang et~al.(2026{\natexlab{b}})Wang, Tan, Zhou, Huang, Zhou, Huang, Ye, Cheng, Qian, Chen, et~al.]{wang2026moss}
Pengyu Wang, Chenkun Tan, Shaojun Zhou, Wei Huang, Qirui Zhou, Zhan Huang, Zhen Ye, Jijun Cheng, Xiaomeng Qian, Yanxin Chen, et~al.
\newblock Moss-video-preview: Toward real-time video understanding via cross-attention.
\newblock \emph{arXiv preprint arXiv:2606.07639}, 2026{\natexlab{b}}.

\bibitem[Wang et~al.(2025{\natexlab{a}})Wang, He, Hong, Cheng, Zhang, Qi, Ding, Gu, Huang, Xu, et~al.]{wang2025lvbench}
Weihan Wang, Zehai He, Wenyi Hong, Yean Cheng, Xiaohan Zhang, Ji~Qi, Ming Ding, Xiaotao Gu, Shiyu Huang, Bin Xu, et~al.
\newblock Lvbench: An extreme long video understanding benchmark.
\newblock In \emph{2025 IEEE/CVF International Conference on Computer Vision (ICCV)}, pp.\  22958--22967. IEEE, 2025{\natexlab{a}}.

\bibitem[Wang et~al.(2025{\natexlab{b}})Wang, Meng, Wang, Zhang, and Zhao]{wang2025proactivevideoqa}
Yueqian Wang, Xiaojun Meng, Yifan Wang, Huishuai Zhang, and Dongyan Zhao.
\newblock Proactivevideoqa: A comprehensive benchmark evaluating proactive interactions in video large language models.
\newblock \emph{arXiv preprint arXiv:2507.09313}, 2025{\natexlab{b}}.

\bibitem[Wang et~al.(2025{\natexlab{c}})Wang, Wang, Chen, Wu, Zhao, and Zheng]{wang2025omnimmi}
Yuxuan Wang, Yueqian Wang, Bo~Chen, Tong Wu, Dongyan Zhao, and Zilong Zheng.
\newblock Omnimmi: A comprehensive multi-modal interaction benchmark in streaming video contexts.
\newblock In \emph{2025 IEEE/CVF Conference on Computer Vision and Pattern Recognition (CVPR)}, pp.\  18925--18935. IEEE, 2025{\natexlab{c}}.

\bibitem[Wu et~al.(2024)Wu, Li, Chen, and Li]{wu2024longvideobench}
Haoning Wu, Dongxu Li, Bei Chen, and Junnan Li.
\newblock Longvideobench: A benchmark for long-context interleaved video-language understanding.
\newblock \emph{Advances in Neural Information Processing Systems}, 37:\penalty0 28828--28857, 2024.

\bibitem[Xu et~al.(2025)Xu, Guo, Hu, Chu, Wang, He, Wang, Shi, He, Zhu, et~al.]{xu2025qwen3}
Jin Xu, Zhifang Guo, Hangrui Hu, Yunfei Chu, Xiong Wang, Jinzheng He, Yuxuan Wang, Xian Shi, Ting He, Xinfa Zhu, et~al.
\newblock Qwen3-omni technical report.
\newblock \emph{arXiv preprint arXiv:2509.17765}, 2025.

\bibitem[Xu et~al.(2026)Xu, Xiao, Chen, He, Peng, Lu, and Han]{xu2026streamingvlm}
Ruyi Xu, Guangxuan Xiao, Yukang Chen, Liuning He, Kelly Peng, Yao Lu, and Song Han.
\newblock Streamingvlm: Real-time understanding for infinite video streams.
\newblock In \emph{International Conference on Learning Representations}, volume 2026, pp.\  61463--61475, 2026.

\bibitem[Yang et~al.(2026)Yang, Zhang, Jia, Guo, Shen, Zhang, Zhang, Wang, Li, Zhang, et~al.]{yang2026mage}
Senqiao Yang, Kaichen Zhang, Zhaoyang Jia, Jinghao Guo, Yifei Shen, Xinjie Zhang, Xiaoyi Zhang, Haoqing Wang, Xiao Li, Peng Zhang, et~al.
\newblock Mage-vl: An efficient codec-native streaming multimodal foundation model.
\newblock \emph{arXiv preprint arXiv:2607.24904}, 2026.

\bibitem[Yao et~al.(2026)Yao, Zhou, Yang, Qin, Hou, Liang, Wang, Cao, Ye, Xie, et~al.]{yao2026joyai}
Dingyu Yao, Junhao Zhou, Chenxu Yang, Chuanyu Qin, Haowen Hou, Zheming Liang, Congcong Wang, Yuhang Cao, Shenglong Ye, Shuai Xie, et~al.
\newblock Joyai-vl-interaction: Real-time vision-language interaction intelligence.
\newblock \emph{arXiv preprint arXiv:2606.14777}, 2026.

\bibitem[Zhang et~al.(2026{\natexlab{a}})Zhang, Chen, Wu, Li, Zhang, Zhang, Liu, Lin, Peng, Liu, et~al.]{zhang2026duplexsla}
Haoyang Zhang, Jun Chen, Donghang Wu, Yuxin Li, Yuxin Zhang, Xiangyu~Tony Zhang, Che Liu, Qingjian Lin, Yizhou Peng, Hexin Liu, et~al.
\newblock Duplexsla: A full-duplex spoken language model with synchronized speech, language, and action.
\newblock \emph{arXiv preprint arXiv:2605.20755}, 2026{\natexlab{a}}.

\bibitem[Zhang et~al.(2026{\natexlab{b}})Zhang, Huang, Wu, Li, and Qi]{zhang2026aero}
Kaichen Zhang, Wei Huang, Keming Wu, Bo~Li, and Xiaojuan Qi.
\newblock Aero realtime: Fully aligned input-output streams for low-latency streaming multimodal generation.
\newblock \emph{arXiv preprint arXiv:2608.08469}, 2026{\natexlab{b}}.

\bibitem[Zhang et~al.(2025)Zhang, Cheng, Deng, Chen, Wang, Zheng, Liu, Yu, Tan, Du, et~al.]{zhang2025omniflatten}
Qinglin Zhang, Luyao Cheng, Chong Deng, Qian Chen, Wen Wang, Siqi Zheng, Jiaqing Liu, Hai Yu, Chao-Hong Tan, Zhihao Du, et~al.
\newblock Omniflatten: An end-to-end gpt model for seamless voice conversation.
\newblock In \emph{Proceedings of the 63rd Annual Meeting of the Association for Computational Linguistics (Volume 1: Long Papers)}, pp.\  14570--14580, 2025.

\bibitem[Zhao et~al.(2025)Zhao, Huang, Zhou, Yao, Ding, Wang, Wang, Liu, and Bin]{zhao2025skim}
Zhengqi Zhao, Xiaohu Huang, Hao Zhou, Kun Yao, Errui Ding, Jingdong Wang, Xinggang Wang, Wenyu Liu, and Feng Bin.
\newblock Skim then focus: Integrating contextual and fine-grained views for repetitive action counting.
\newblock \emph{International Journal of Computer Vision}, 133\penalty0 (9):\penalty0 6347--6361, 2025.

\bibitem[Zhou et~al.(2025)Zhou, Shu, Zhao, Wu, Liang, Xiao, Qin, Yang, Xiong, Zhang, et~al.]{zhou2025mlvu}
Junjie Zhou, Yan Shu, Bo~Zhao, Boya Wu, Zhengyang Liang, Shitao Xiao, Minghao Qin, Xi~Yang, Yongping Xiong, Bo~Zhang, et~al.
\newblock Mlvu: Benchmarking multi-task long video understanding.
\newblock In \emph{2025 IEEE/CVF Conference on Computer Vision and Pattern Recognition (CVPR)}, pp.\  13691--13701. IEEE, 2025.

\end{thebibliography}
\bibliographystyle{iclr2027_conference}

\newpage
\appendix

\section{Details of LLM as a judge}
\label{app:evaluation_details}

\subsection{Model Access and Deployment}

We evaluate the cloud-based models, including
GPT-Realtime-2.1\footnote{\href{https://developers.openai.com/api/docs/models/gpt-realtime-2.1}{OpenAI API}},
Gemini-3.1-Flash-Live\footnote{\href{https://ai.google.dev/gemini-api/docs/models/gemini-3.1-flash-live-preview}{Google Gemini API}},
and Qwen3.5-Omni-Plus-Realtime\footnote{\href{https://www.alibabacloud.com/help/en/model-studio/qwen3-5-omni-plus-realtime}{Alibaba Cloud Model Studio}},
through their respective providers' official platforms.

\subsection{Streaming Input Protocol}

For native audio-visual models, we preserve the original audio and visual
content and its temporal ordering, supplying these inputs through each
model's supported interface. For streaming visual models, we retain the
video and replace the audio with timestamp-aligned transcripts. Transcript
content is revealed incrementally according to the original speech timeline,
so that future utterances are not exposed in advance. The underlying videos,
task instructions, and temporal references remain the same across the two
evaluation settings.

\subsection{Evaluation Prompts}
\label{app:evaluation_prompts}

Following the evaluation protocol in the main text, we use GPT-5.6 Terra
as the judge for binary response-quality evaluation. The task-specific judge prompts used in our experiments are provided below. Fields enclosed in braces are populated with instance-specific inputs.


\begin{divepromptbox}{Task1 VTI --- Procedural Intervention}
Please determine whether the model's output correctly completes the
intervention task at the annotated event.

\medskip
\noindent\textbf{Task Description:}\par
A first-person procedure contains an annotated operation error. Judge whether
the reply performs a useful intervention on that error. The user's instruction
need not be wrong; the observed operation described by the reference may be
wrong. This benchmark has one reference marker per Task1 case, so this is one
point-level decision.

\medskip
\noindent\textbf{Input:}\par
\noindent User instruction: \texttt{\{instruction\}}\par
\noindent Ground-truth correction: \texttt{\{ground\_answer\}}\par
\noindent Reference annotation and context:
\texttt{\{reference\_annotation\}}\par
\noindent Model output selected for this event:
\texttt{"\{response\}"}

\medskip
\noindent\textbf{Criteria:}
\begin{enumerate}
    \setlength{\itemsep}{0pt}
    \setlength{\parsep}{0pt}
    \item \textbf{Core-error identification:} The reply must identify the
    core operation error from the instruction and reference, or give a
    correction that makes that error unambiguous. Generic encouragement,
    vague caution, or correcting only another problem fails.

    \item \textbf{Context consistency:} Preserve the correct action,
    object/tool, and necessary constraints. A concise correction such as
    ``use the butter knife instead'' suffices when the context makes the
    intended problem and correction clear.

    \item \textbf{Essential correction coverage:} Accept reasonable
    equivalent remedies and omit optional explanations or secondary details.
    If several corrections are necessary to resolve the core problem, all
    those necessary corrections must be covered.

    \item \textbf{Extra-information handling:} Harmless additional
    information does not affect success. Approval of the erroneous operation,
    advice contradicting the required correction, or advice that would
    continue the core error is a failure.

    \item \textbf{Non-answer handling:} When the reference states the
    operation error, merely requesting the procedure, asking for more
    observations, or claiming no procedure was provided fails to intervene
    and receives 0. This does not make the reference unassessable.
\end{enumerate}

\noindent\textbf{Scoring:}\par
\noindent 1 = The core operation error is clearly identified/corrected with
a useful intervention.\par
\noindent 0 = The core error is missed, the intervention is materially
wrong/incomplete, or no reply is provided.

\medskip
\noindent\textbf{Output Format:}\par
Only output a JSON object in the following format, replacing the
placeholders with the score and explanation:
\begin{quote}
\ttfamily\small
\{\par
\hspace*{1em}"success\_score": <0 or 1>,\par
\hspace*{1em}"reasoning": "<explanation>"\par
\}
\end{quote}
\end{divepromptbox}

\begin{divepromptbox}{Task2 SPI --- Proactive Warning}
Please determine whether the model's output correctly completes the warning
or response task at each annotated event.

\medskip
\noindent\textbf{Task Description:}\par
An audio-visual conversation progressively reveals risk. At each reference
marker, evaluate the warning or conversational response required at that
particular point. Some reference turns acknowledge a safe plan or answer an
ordinary request; do not invent a danger that the current reference does not
require the model to warn about.

\medskip
\noindent\textbf{Input:}\par
\noindent User instruction and current request:
\texttt{\{instruction\}}\par
\noindent Ground-truth warning or response:
\texttt{\{ground\_answer\}}\par
\noindent Reference annotation and prior dialogue:
\texttt{\{reference\_annotation\}}\par
\noindent Model output selected for this event:
\texttt{"\{response\}"}

\medskip
\noindent\textbf{Criteria:}
\begin{enumerate}
    \setlength{\itemsep}{0pt}
    \setlength{\parsep}{0pt}

    \item \textbf{Required response type:} Use the current reference and
    prior context to determine whether this point requires a risk warning,
    an ordinary answer, or a safe-plan acknowledgment. A brief relevant
    answer or acknowledgment can satisfy the latter two types; do not invent
    a danger that the reference does not require.

    \item \textbf{Risk-specific intervention:} At a risk point, identify the
    relevant danger or provide a specific safer action that unambiguously
    addresses it. ``Do not mix those cleaners'' can suffice without a
    chemistry explanation. Vague concern or a warning about only another
    danger fails to address the required risk.

    \item \textbf{Scenario consistency:} Preserve the essential people,
    action, and circumstances. Do not confuse a proposed risky action with
    an actual accident when that changes the appropriate advice.

    \item \textbf{Safety requirements:} Accept alternative safe advice and
    concise warnings without optional examples or secondary explanation.
    Safety-critical conditions must still be covered; approving the
    dangerous action or giving materially contradictory or unsafe advice is
    a failure.
\end{enumerate}

\noindent\textbf{Scoring:}\par
\noindent 1 = Completes this point's core warning or conversational
purpose.\par
\noindent 0 = Misses or misidentifies that purpose, gives materially
unsafe/inconsistent advice, or provides no reply.

\medskip
\noindent\textbf{Output Format:}\par
Only output a JSON object in the following format, replacing the
placeholders with the success sequence and explanation:
\begin{quote}
\ttfamily\small
\{\par
\hspace*{1em}"success\_sequence":
"<01 sequence in input marker order>",\par
\hspace*{1em}"reasoning": "<explanation>"\par
\}
\end{quote}
\end{divepromptbox}

\begin{divepromptbox}{Task3 CQI --- Continuous Query Interaction}
Please determine whether the model's output correctly answers the current
question at each annotated event.

\medskip
\noindent\textbf{Task Description:}\par
A continuous interaction contains successive questions, whose speech may
overlap. At each reference marker, evaluate the answer to that marker's
question using the applicable instruction, reference answer, and supplied
context. Questions may depend on visual evidence or be independent of it;
use the supplied annotations to determine what is required. Timing and
output-to-marker association are handled by the program.

\medskip
\noindent\textbf{Input:}\par
\noindent Current user question or interaction instruction:
\texttt{\{instruction\}}\par
\noindent Ground-truth answer to the current question:
\texttt{\{ground\_answer\}}\par
\noindent Reference annotation, available visual context, and prior dialogue:
\texttt{\{reference\_annotation\}}\par
\noindent Model output selected for this event:
\texttt{"\{response\}"}

\medskip
\noindent\textbf{Criteria:}
\begin{enumerate}
    \setlength{\itemsep}{0pt}
    \setlength{\parsep}{0pt}

    \item \textbf{Current-question alignment:} Answer the question assigned
    to this marker. A correct answer to an earlier, later, or different
    question does not satisfy the current one. Use prior dialogue only where
    needed to resolve references or an applicable request; do not import
    later questions, updates, or answers into the current event.

    \item \textbf{Answer fulfillment:} Provide the essential answer requested
    at this point. Accept semantically equivalent wording, concise answers,
    and valid alternatives supported by the supplied reference. A name,
    number, or short phrase can suffice when it fully answers the question.
    Mere acknowledgment, repetition of the question, or a promise to answer
    is insufficient when a substantive answer is required.

    \item \textbf{Contextual grounding:} For a visually grounded question,
    preserve the essential objects, actions, attributes, counts, or relations
    established by the reference and the applicable scene annotations. Do
    not accept a generic plausible answer that conflicts with them. For a
    video-independent question, evaluate its requested factual or reasoning
    content without imposing an unnecessary visual requirement. Do not infer
    unseen visual evidence from the candidate's own claims.

    \item \textbf{Completeness and consistency:} Satisfy all essential parts
    of the current request without requiring optional detail or an
    explanation that was not requested. An explicit self-correction may
    replace an earlier mistake when the final intended answer is clear. An
    unresolved contradiction or a materially incorrect claim about the
    requested answer is a failure.

    \item \textbf{Reference limits:} Missing question transcription, raw
    media, or an optional context field alone does not block judgment when
    the supplied reference establishes the required answer. If essential
    question content, reference-answer information, or contextual facts are
    unavailable and prevent a judgment, apply the shared missing-reference
    exception. Do not reconstruct the question or reference answer from the
    candidate output, timestamps, or audio-file names. An incorrect, vague,
    or absent candidate answer is 0 when the reference is sufficient.
\end{enumerate}

\noindent\textbf{Scoring:}\par
\noindent 1 = Correctly answers the current question and satisfies its
essential reference-supported requirements.\par
\noindent 0 = Answers the wrong question, omits or contradicts essential
answer content, fails the request, or provides no reply.

\medskip
\noindent\textbf{Output Format:}\par
Only output a JSON object in the following format, replacing the
placeholders with the success sequence and explanation:
\begin{quote}
\ttfamily\small
\{\par
\hspace*{1em}"success\_sequence":
"<01 sequence in input marker order>",\par
\hspace*{1em}"reasoning": "<explanation>"\par
\}
\end{quote}
\end{divepromptbox}

\begin{divepromptbox}{Task4 ORA --- Response Adaptation}
Please determine whether the model's output correctly adapts its response
to the applicable request or update at each annotated event.

\medskip
\noindent\textbf{Task Description:}\par
Evaluate assistant-audio cases only. A dialogue may be interrupted by a
corrected fact, new constraint, or changed request. At each reference point,
judge the actual response requirement and any update already applicable at
that time.

\medskip
\noindent\textbf{Input:}\par
\noindent Current user instruction or request:
\texttt{\{instruction\}}\par
\noindent Ground-truth updated answer:
\texttt{\{ground\_answer\}}\par
\noindent Reference annotation, prior dialogue, and applicable interruption:
\texttt{\{reference\_annotation\}}\par
\noindent Model output selected for this event:
\texttt{"\{response\}"}

\medskip
\noindent\textbf{Criteria:}
\begin{enumerate}
    \setlength{\itemsep}{0pt}
    \setlength{\parsep}{0pt}

    \item \textbf{Applicable update:} Identify the request or update already
    in effect at this point, preserving its subject, object, and scope. Do
    not apply a later update to an earlier reply. Ordinary turns without an
    update use their own reference-supported request.

    \item \textbf{Answer fulfillment:} A short revised answer succeeds if
    it incorporates the essential update and answers the core request. Mere
    ``understood'' is insufficient when a substantive answer is required.
    Announcing the interruption, explaining the revision, or repeating the
    full old answer is unnecessary.

    \item \textbf{Self-correction and consistency:} An explicit correction
    may replace an earlier mistake when the final intended answer is clear.
    Continuing to use a superseded fact or constraint, or leaving
    contradictory advice unresolved, is a failure.

    \item \textbf{Reference limits:} Do not require optional details or a
    continuation absent from an interrupted reference. Missing user
    transcription alone does not block judgment when the reference
    establishes the requirement; never invent missing audio content or an
    update.
\end{enumerate}

\noindent\textbf{Scoring:}\par
\noindent 1 = Incorporates the applicable update and fulfills the point's
core answer requirement.\par
\noindent 0 = Ignores or misapplies an essential update, fails the request,
or provides no reply.

\medskip
\noindent\textbf{Output Format:}\par
Only output a JSON object in the following format, replacing the
placeholders with the success sequence and explanation:
\begin{quote}
\ttfamily\small
\{\par
\hspace*{1em}"success\_sequence":
"<01 sequence in input marker order>",\par
\hspace*{1em}"reasoning": "<explanation>"\par
\}
\end{quote}
\end{divepromptbox}
\begin{divepromptbox}{Task5 AVAT --- Per-Event Response Quality}
Please determine whether the first model response assigned to the current
response-required event fulfills the conversational act and content
requirements of that event.

\medskip
\noindent\textbf{Task Description:}\par
Evaluate only the first model response assigned to the current
response-required event, using that event's reference information and
applicable conversational context. A response succeeds only if it fulfills
the current event's required conversational act and essential content
requirements.

\medskip
\noindent\textbf{Input:}\par
\noindent User instruction or scene description:
\texttt{\{instruction\}}\par
\noindent Current event reference answer:
\texttt{\{ground\_answer\}}\par
\noindent Current event reference, including marker number (one-based),
reference window in seconds, transcription, participant ID, and target of
speech:
\texttt{\{reference\_annotation\}}\par
\noindent First model response assigned to this event:
\texttt{"\{response\}"}

\medskip
\noindent\textbf{Criteria:}
\begin{enumerate}
\setlength{\itemsep}{0pt}
\setlength{\parsep}{0pt}

\item \textbf{Event-level fulfillment:} Give 1 only if the assigned
response fulfills the current event's required conversational act and
essential content requirements, according to the event reference and
applicable context.

\item \textbf{Content grounding:} The response must address the current
event itself. Merely quoting or summarizing another participant's
utterance, or making a correct observation about the conversation, is
insufficient unless the current request explicitly asks for that content
or analysis.

\item \textbf{Semantic equivalence:} Accept equivalent wording,
paraphrases, and harmless additional details when the required meaning
is preserved. A material contradiction concerning the required answer
results in 0.

\item \textbf{Response scope:} Evaluate only the first response assigned
to the current event. Do not use another response elsewhere in the case
to compensate for an incorrect or missing response at this event. A
missing assigned response receives 0.

\item \textbf{Separate dimensions:} Response timing and speak/silence
decisions are evaluated separately. Do not include whether the response
was early, on-time, or late, or whether other events were answered or
left silent, in this content-quality judgment.

\item \textbf{Judgment evidence:} Base the judgment only on the supplied
instruction, current-event reference, applicable context, and assigned
response. 

\end{enumerate}

\noindent\textbf{Scoring:}\par
\noindent 1 = The assigned response fulfills the current event's required
conversational act and essential content requirements.\par
\noindent 0 = The response fails to fulfill the current event, materially
contradicts the required answer, or no response is assigned.

\medskip
\noindent\textbf{Output Format:}\par
Only output a JSON object in the following format, replacing the
placeholders with the score and explanation:
\begin{quote}
\ttfamily\small
\{\par
\hspace*{1em}"success\_score": <0 or 1>,\par
\hspace*{1em}"reasoning": "<explanation>"\par
\}
\end{quote}
\end{divepromptbox}

\begin{divepromptbox}{Task6 CAT --- Per-Marker Action Count}
Please determine whether the model's output correctly reports the action
count at each annotated event.

\medskip
\noindent\textbf{Task Description:}\par
Each completed target action has its own reference count marker. Judge only
the reply assigned by the benchmark to this marker and selected by the
program. A final total is not a substitute for earlier counting responses.
Missing points are already scored 0.

\medskip
\noindent\textbf{Input:}\par
\noindent Action-counting instruction:
\texttt{\{instruction\}}\par
\noindent Expected count at this marker:
\texttt{\{ground\_answer\}}\par
\noindent Reference marker and task context:
\texttt{\{reference\_annotation\}}\par
\noindent Model output assigned to this marker:
\texttt{"\{response\}"}

\medskip
\noindent\textbf{Criteria:}
\begin{enumerate}
    \setlength{\itemsep}{0pt}
    \setlength{\parsep}{0pt}

    \item \textbf{Count grounding:} Determine the instructed action and
    expected count at this marker. The reply must report that action count,
    not merely mention the number for another purpose.

    \item \textbf{Accepted expression:} Accept digits, equivalent number
    words, and brief natural wording. The correct count alone suffices; an
    action description or explanation is unnecessary, and harmless filler
    does not affect the decision.

    \item \textbf{Count disambiguation:} A wrong number, a neighboring count
    alone, or guessing several alternatives fails. When multiple numbers
    appear, the intended count for this marker must be unambiguous; merely
    including the expected number in a list is insufficient.

    \item \textbf{Self-correction:} An explicit correction succeeds when it
    clearly replaces an incorrect count with the expected one. Unresolved
    conflicting counts for this marker fail.

\end{enumerate}

\noindent\textbf{Scoring:}\par
\noindent 1 = Clearly reports the expected count for this assigned marker.\par
\noindent 0 = Missing, wrong, unrelated, or ambiguous count, or no reply at
this marker.

\medskip
\noindent\textbf{Output Format:}\par
Only output a JSON object in the following format, replacing the
placeholders with the success sequence and explanation:
\begin{quote}
\ttfamily\small
\{\par
\hspace*{1em}"success\_sequence":
"<01 sequence in input marker order>",\par
\hspace*{1em}"reasoning": "<explanation>"\par
\}
\end{quote}
\end{divepromptbox}

\section{Human Annotation Details}
\label{app:human_annotation}

\subsection{Annotation Procedure}

All human annotations in DIVE were obtained by consolidating annotations
from four independent annotators. The temporal annotations cover both
the earliest valid response points and task-specific response windows.
The former identify when a response first becomes appropriate, while
the latter specify the intervals during which the model is expected to
respond.

\subsection{Earliest Response Points}

For Continuous Action Tracking (CAT), each annotator identifies the
timestamp at which a target action is completed and the corresponding
count can first be reported. The final reference timestamp is the
arithmetic mean of the four annotations:
\begin{equation}
    t_j^{\mathrm{ref}} = \frac{1}{4}\sum_{k=1}^{4} t_j^{(k)},
    \label{eq:cat_annotation_mean}
\end{equation}
where $t_j^{(k)}$ is the timestamp assigned by annotator $k$ to the
$j$-th completed action. This timestamp serves as the earliest response
reference for the corresponding counting event.

For Video Timely Intervention (VTI), annotators also identify the earliest
moment at which the error becomes visually recognizable, providing the
reference point for evaluating intervention latency.

\subsection{Task-Specific Response Windows}

Following the notation in the main text, we denote a valid response
window by $
W_j=[t_j^{\mathrm{start}},t_j^{\mathrm{end}}).$.
Its boundaries are determined by the interaction requirements of each task:

\textbf{Video Timely Intervention (VTI):}
    The response window extends from the occurrence of an error to
    the completion of the corresponding erroneous action.

\textbf{Speech Proactive Intervention (SPI):}
    For scenarios involving auditory or visual cues, the response window
    is the gap between two relevant consecutive utterances by the speaker
    in the video. It begins when the first utterance ends and closes when
    the next utterance begins.

\textbf{Overlapping Response Adaptation (ORA):}
    The response window begins when the information requiring a response
    first becomes fully available in the video and ends when the person
    in the video initiates their own reaction.

 \textbf{Audio-Visual Addressee Tracking (AVAT):}
    For slots requiring a model response, the response window is the gap
    between utterances by the other participants, from the end of one
    utterance to the onset of the next. Conversational context and visual
    addressee cues determine whether the model should speak within that gap.

\textbf{Continuous Query Interaction (CQI):}
    The response window for a question begins when that question ends and
    closes when the subsequent question ends. Specifically, if $e_j$
    denotes the end time of question $j$, the window for its response is
    $W_j=[e_j,e_{j+1})$.


\end{document}